\documentclass[journal]{IEEEtran}
\usepackage{amsmath,amssymb,bm}
\usepackage{booktabs}
\usepackage{multirow}
\usepackage{graphicx}
\usepackage{xcolor}
\usepackage{url}
\usepackage{cite}
\usepackage{algorithm}
\usepackage{algorithmic}
\usepackage{array}
\usepackage{balance}

\newcommand{\method}{\textsc{BiSCo-LLM}}

\newcommand{\R}{\mathbb{R}}
\newcommand{\sign}{\operatorname{sign}}
\newcommand{\sg}{\operatorname{sg}}

\newcommand{\Assemble}{\operatorname{Assemble}}

\newcommand{\TopK}{\operatorname{TopK}}

\begin{document}

\title{BiSCo-LLM: Lookup-Free Binary Spherical Coding for Extreme Low-Bit Large Language Model Compression}

\author{Yuantian~Shao\textsuperscript{1,2,\dag},
Peisong~Wang\textsuperscript{2,\dag,*},
Zhilei~Liu\textsuperscript{2},
Chuangyi~Li\textsuperscript{2},
Yuanteng~Chen\textsuperscript{2},
Pengcheng~Xie\textsuperscript{3},
Yiwu~Yao\textsuperscript{3},
Zhihui~Wei\textsuperscript{1,*},
and Jian~Cheng\textsuperscript{2,*}%
\thanks{\textsuperscript{\dag}Yuantian Shao and Peisong Wang contributed equally to this work.}%
\thanks{\textsuperscript{*}Peisong Wang, Zhihui Wei, and Jian Cheng are corresponding authors.}%
\thanks{\textsuperscript{1}Yuantian Shao and Zhihui Wei are with Nanjing University of Science and Technology, Nanjing, China.}%
\thanks{\textsuperscript{2}Yuantian Shao, Peisong Wang, Zhilei Liu, Chuangyi Li, Yuanteng Chen, and Jian Cheng are with the Institute of Automation, Chinese Academy of Sciences, Beijing, China.}%
\thanks{\textsuperscript{3}Pengcheng Xie and Yiwu Yao are with Huawei, China.}%
% \thanks{This work was supported in part by \todo{funding information}.}%
}

%\markboth{IEEE Transactions on Neural Networks and Learning Systems,~Vol.~XX, No.~XX, 2026}%
%{First Author \MakeLowercase{\textit{et al.}}: BiSCo-LLM}

\maketitle

\begin{abstract}
Large language models (LLMs) are increasingly constrained by memory capacity, weight bandwidth, and checkpoint storage during deployment. Existing low-bit compression methods mainly follow two directions. Scalar or group-wise quantization is simple and compatible with efficient low-precision kernels, but its representation capacity becomes limited when the target budget approaches 2 bits per weight. Vector-quantized weight compression provides a richer block-level representation, but usually introduces explicit codebooks, index lookup, and additional storage accounting. This paper presents \method{}, a codebook-free binary spherical coding framework for extreme low-bit LLM weight compression. The proposed pipeline is built on three components. First, local weight chunks are mapped onto a unit hypersphere and binarized into compact spherical codes, so that the main payload is a bit-packed sign stream rather than explicit VQ centroids. Second, a residual BSQ stage encodes the reconstruction error left by the base spherical codec, providing an explicit rate--distortion path without stored codebooks. Third, category-wise recovery distillation is performed after replacing each Transformer module category, reducing the mismatch between local weight reconstruction and assembled model behavior. A small 8-bit protected-channel path is used as an auxiliary stabilization mechanism for sensitive channels and is counted separately from the BSQ payload. The reported storage budget includes binary codes, neural decoders, protected-channel payloads, LoRA adapters, and metadata. On Qwen3-8B, \method{} obtains a WikiText-2 perplexity of 10.18 compared with 9.73 for the FP16/BF16 model, and an average downstream accuracy of 68.05 compared with 69.92 over the reported seven-task evaluation set. These results indicate that codebook-free spherical coding can preserve model behavior under an extreme low-bit storage budget when residual coding, sensitivity-aware protection, and recovery distillation are jointly considered.
% 中文翻译：大语言模型在部署时越来越受到显存容量、权重带宽和 checkpoint 存储的限制。现有低比特压缩方法主要有两个方向。标量量化或分组量化简单，并且兼容高效低精度 kernel，但当目标预算接近每个权重 2 bit 时，其表示能力会受到限制。向量量化权重压缩提供了更强的块级表示能力，但通常会引入显式码本、索引查找以及额外的存储开销核算。本文提出 BiSCo-LLM，一种面向极低比特 LLM 权重压缩的无码本二值球面编码框架。所提出流程由三个组件构成。第一，局部权重块被映射到单位超球面上，并被二值化为紧凑的球面码，使主要载荷成为打包后的符号码流，而不是显式 VQ 中心向量。第二，残差 BSQ 阶段编码基础球面 codec 留下的重建误差，在不存储码本的情况下提供明确的率失真路径。第三，每替换一个 Transformer 模块类别后执行逐类别恢复蒸馏，以降低局部权重重建与组装后模型行为之间的错配。少量 8bit 保护通道路径被用作敏感通道的辅助稳定机制，并与 BSQ 主载荷分开统计。报告的存储预算包括二值码、神经解码器、保护通道载荷、LoRA adapter 和元数据。在 Qwen3-8B 上，BiSCo-LLM 的 WikiText-2 困惑度为 10.18，而 FP16/BF16 模型为 9.73；在报告的七任务评估集合上的平均下游精度为 68.05，而 FP16/BF16 模型为 69.92。这些结果表明，在联合考虑残差编码、敏感性感知保护和恢复蒸馏时，无码本球面编码可以在极低比特存储预算下保持模型行为。
\end{abstract}

\begin{IEEEkeywords}
Large language model compression, low-bit quantization, binary spherical coding, codebook-free compression, vector quantization, residual coding, LoRA compensation.
\end{IEEEkeywords}

\section{Introduction}
Large language models (LLMs) have become foundation components for language understanding, reasoning, generation, and general-purpose AI services \cite{brown2020language,touvron2023llama,achiam2023gpt4}. Despite their effectiveness, the continuous growth of model scale has introduced substantial deployment challenges. Recent technical reports show that LLM deployment targets have moved from several-billion-parameter models to hundreds-of-billions and trillion-scale architectures. DeepSeek-V3 and DeepSeek-R1, for example, adopt a 671B-parameter MoE architecture with 37B activated parameters per token; Kimi K2 scales this regime to a 1T-parameter MoE model with 32B activated parameters; and GLM-4.5 reports 355B total parameters with 32B activated parameters \cite{liu2024v3,guo2025r1,kimi2025k2,zeng2025arc}. These models correspond to terabyte-scale BF16/FP16 checkpoint storage before KV caches, runtime buffers, and serving-system overhead are considered. As a result, efficient inference is constrained not only by peak arithmetic throughput, but also by the amount of model data that must be fetched from memory and moved through the memory hierarchy \cite{frantar2022gptq,sheng2023flexgen}. This constraint becomes particularly pronounced during autoregressive decoding with limited batch sizes, where each generated token requires repeatedly applying large linear transformations and the available computation may be insufficient to amortize weight-loading cost. In this regime, weight movement from high-bandwidth memory can dominate latency and throughput \cite{kim2023finequant,lin2023awq}. Reducing the stored precision of model weights to 4 bits, 3 bits, 2 bits, or below is therefore a direct and system-relevant approach to decreasing memory footprint and relieving bandwidth pressure.
% 中文翻译：大语言模型已经成为语言理解、推理、生成以及通用 AI 服务的基础组件。尽管大语言模型表现出很强的能力，但模型规模的持续增长给实际部署带来了显著挑战。近期技术报告表明，LLM 的部署对象已经从数十亿参数模型扩展到数千亿乃至万亿参数架构。例如，DeepSeek-V3 和 DeepSeek-R1 采用 671B 总参数、每 token 激活 37B 参数的 MoE 架构；Kimi K2 将这一规模进一步扩展到 1T 总参数、32B 激活参数的 MoE 模型；GLM-4.5 报告了 355B 总参数和 32B 激活参数。这些模型在 BF16/FP16 表示下对应 TB 级 checkpoint 存储，并且尚未计入 KV cache、运行时 buffer 和服务系统开销。因此，高效推理不仅受峰值算术吞吐限制，也受到需要从显存中读取并在存储层级中搬运的模型数据量限制。这一约束在 batch size 受限的自回归解码中尤其明显，因为每生成一个 token 都需要反复执行大规模线性变换，而有限的 batch size 不足以充分摊销权重加载成本。在这种情况下，从高带宽显存中搬运权重可能主导推理延迟和吞吐。因此，将模型权重的存储精度压缩到 4 bit、3 bit、2 bit 甚至更低，是降低显存占用并缓解带宽压力的一种直接且具有系统意义的方法。

The demand for compression is further reinforced by recent changes in LLM deployment scenarios. On-device and hybrid personal assistants require low-latency inference under strict memory, energy, and privacy constraints, as exemplified by Apple Intelligence, which combines an on-device foundation model with a larger server-side model for more complex requests \cite{gunter2024apple}. In parallel, agentic and coding-oriented systems convert a user request into a multi-step inference process involving planning, tool invocation, code execution, verification, and iterative refinement \cite{rama2025cerebrum,wang2025openhands}. Such workflows can substantially increase the number of model invocations required to complete a single task. In enterprise, private-cloud, and edge deployments, model replicas may also be distributed across multiple tenants, accelerators, and locations, making storage footprint, memory bandwidth, and serving density system-level constraints. These trends require compression methods that reduce deployable model size while preserving task-level behavior.
% 中文翻译：近期 LLM 部署场景的变化进一步强化了模型压缩需求。端侧和混合式个人助手需要在严格的存储、能耗和隐私约束下实现低延迟推理，例如 Apple Intelligence 同时使用端侧基础模型和更大的服务端模型来处理不同复杂度的请求。与此同时，面向智能体和代码任务的系统会将单个用户请求转化为包含规划、工具调用、代码执行、验证和迭代修正的多步推理过程。这类工作流会显著增加完成单个任务所需的模型调用次数。在企业、私有云和边缘部署中，模型副本还可能分布在多个租户、加速器和地理位置上，使存储占用、显存带宽和服务密度成为系统级约束。这些趋势要求压缩方法不仅降低可部署模型大小，还要保持任务级行为。

Post-training quantization (PTQ) has become a central approach for addressing these deployment constraints, since it reduces the numerical precision of pretrained models without requiring full retraining. Existing PTQ methods have improved LLM compression from several complementary perspectives. Early systems such as ZeroQuant combine layer-wise distillation with system-aware kernels, demonstrating the feasibility of quantizing large Transformer models after pretraining \cite{yao2022zeroquant}. GPTQ reduces weight reconstruction error through a layer-wise second-order formulation that uses calibration activations to approximate the local Hessian of the reconstruction objective \cite{frantar2022gptq}. OmniQuant follows a different calibration-based optimization route by learning clipping thresholds, quantization scales, and equivalent transformations from calibration data \cite{shao2023omniquant}. Another line of work addresses the pronounced outlier and sensitivity imbalance in LLMs: LLM.int8() separates emergent outlier features through mixed-precision decomposition; RPTQ reorders channels to reduce activation range imbalance; SmoothQuant transfers activation difficulty to weights via equivalent scaling; AWQ protects activation-salient channels; and SpQR and SqueezeLLM use sparse outlier storage or dense-and-sparse decompositions for near-lossless low-bit compression \cite{dettmers2022llmint8,yuan2023rptq,xiao2022smoothquant,lin2023awq,dettmers2023spqr,kim2023squeezellm}. More system-oriented studies, such as Atom, QuaRot, and SpinQuant, further consider serving-friendly mixed precision and rotation-based outlier suppression \cite{zhao2023atom,ashkboos2024quarot,liu2024spinquant}.
% 中文翻译：后训练量化已经成为缓解上述部署约束的核心技术路线，因为它可以在不进行完整重训练的情况下降低预训练模型的数值精度。现有 PTQ 方法从多个互补角度改进了 LLM 压缩。ZeroQuant 等早期系统结合逐层蒸馏和系统感知 kernel，证明了大规模 Transformer 可以在预训练后进行量化。GPTQ 通过逐层二阶优化形式降低权重重建误差，其中校准激活被用于近似重建目标的局部 Hessian。OmniQuant 则采用另一种基于校准数据的优化路线，通过学习截断阈值、量化尺度和等价变换来改善量化质量。另一类工作关注 LLM 中显著的离群值和敏感性不均衡问题：LLM.int8() 通过混合精度分解分离突现离群特征；RPTQ 通过通道重排降低激活范围不均衡；SmoothQuant 通过等价缩放将激活侧困难转移到权重侧；AWQ 保护激活显著通道；SpQR 和 SqueezeLLM 则通过稀疏离群值存储或 dense-and-sparse 分解实现近乎无损的低比特压缩。Atom、QuaRot 和 SpinQuant 等更偏系统的方法进一步考虑面向服务的混合精度和基于旋转的离群值抑制。

Despite these advances, extending LLM quantization to 2 bits or lower remains substantially more challenging than conventional 4-bit or 3-bit settings \cite{zhu2025squeeze10,li2025icquant}. Most PTQ methods still represent weights using scalar or group-wise quantized values. As the bit budget decreases, these representations provide only a small number of reconstruction levels within each group, making it difficult to preserve both the dominant structure of local weight vectors and the directions that are important for maintaining layer outputs. Recent sub-2-bit and 2--3-bit studies support this view: Squeeze10-LLM relies on staged mixed precision and activation-level supervision rather than uniform scalar reconstruction, while ICQuant explicitly models outlier statistics to reduce range expansion and bit overhead in extreme compression regimes \cite{zhu2025squeeze10,li2025icquant}. These results seem to suggest that the problem of extremely low bit compression is difficult to fully solve in scalar quantization. Consequently, recent studies have increasingly explored vector quantization or structured discrete representations \cite{zhang2025glvq,ouderaa2026llvq}. 
% 中文翻译：尽管上述方法已经取得了显著进展，但将 LLM 量化进一步推进到 2 bit 或更低仍然明显难于常规的 4 bit 或 3 bit 设置。大多数 PTQ 方法仍然使用标量量化或分组量化数值来表示权重。随着比特预算降低，这类表示在每个权重组内只能提供很少的重建取值，因此难以同时保留局部权重向量的主要结构，以及那些对维持层输出具有重要作用的方向。近期 sub-2-bit 和 2--3-bit 研究也支持这一观点：Squeeze10-LLM 依赖分阶段混合精度和激活级监督，而不是均匀的标量重建；ICQuant 则显式建模离群值统计信息，以在极低比特压缩中降低量化范围膨胀和比特开销。这些结果表明，在近似均匀的标量比特分配下，仅最小化平均重建误差不足以解决极低比特压缩问题。因此，近期研究越来越多地转向向量化或结构化离散表示。

Vector quantization (VQ) and structured discrete representations provide a stronger representation. Instead of rounding each weight independently, these methods encode a group of weights as a structured discrete object. In recent LLM compression, this direction has evolved from explicit multi-codebook representations to more optimization- and structure-aware formulations. AQLM adapts additive quantization to LLM weights by representing a weight vector as the sum of multiple learned codewords \cite{egiazarian2024aqlm}. VPTQ further formulates vector PTQ with second-order information and introduces residual and outlier quantization to improve extreme low-bit approximation \cite{liu2024vptq}. PocketLLM moves toward learned latent-space weight compression, where pretrained weights are encoded into discrete latent vectors and reconstructed by a lightweight meta-network decoder \cite{tian2025pocketllm}. More recent methods reduce the rigidity or overhead of explicit VQ in different ways: LiftQuant obtains quasi-continuous bit-width control through dimensional lifting and projection from a high-dimensional 1-bit lattice, UniSVQ parameterizes codewords as affine transforms of integer lattices to bridge scalar and vector quantization, and LC-QAT uses linear-constrained VQ to make 2-bit vector quantization compatible with data-efficient QAT \cite{he2026liftquant,wang2026unisvq,wang2026lcqat}. Overall, these methods indicate that the main difficulty of extreme compression is not only the number of scalar levels, but also how local weight vectors are parameterized, optimized, and reconstructed under a strict storage budget.
% 中文翻译：向量量化和结构化离散表示提供了更强的表示能力。它们不是独立舍入每个权重，而是将一组权重编码为一种结构化离散对象。在近期 LLM 压缩中，这一路线已经从显式多码本表示，发展到更重视优化形式和结构约束的方案。AQLM 将加性量化用于 LLM 权重，用多个学习码字之和表示一个权重向量。VPTQ 进一步将向量后训练量化表述为带二阶信息的优化问题，并引入残差量化和离群值量化来改善极低比特近似。PocketLLM 则转向可学习潜空间权重压缩，将预训练权重编码为离散潜变量，并通过轻量级 meta-network 解码器重建。更新近的方法从不同角度降低显式 VQ 的刚性或开销：LiftQuant 通过维度提升以及从高维 1-bit lattice 到原始空间的投影实现准连续比特宽度控制；UniSVQ 将码字参数化为整数 lattice 的仿射变换，以连接标量量化和向量量化；LC-QAT 使用线性约束 VQ，使 2-bit 向量量化能够与数据高效的 QAT 兼容。总体而言，这些方法表明，极低比特压缩的主要困难不仅在于标量取值数量不足，也在于如何在严格存储预算下参数化、优化和重建局部权重向量。

The above progression suggests that vector-level coding is promising, but it also leaves open how a binary latent codec should be optimized for pretrained LLM weights. Lookup-free binary quantization removes the explicit codebook from storage, but unconstrained binary latents may still entangle code direction and latent magnitude during training. BSQ addresses this issue in visual tokenization by normalizing latent vectors onto a hypersphere before binarization, so that the discrete code mainly represents direction while the decoder receives a scale-controlled binary input \cite{yu2024magvitv2,zhao2024bsq}. This property is particularly relevant to weight compression: instead of mapping each weight chunk to a fixed lattice point or an explicit centroid, a spherical binary code can provide a compact codebook-free payload, while a shared decoder learns the reconstruction rule for a family of weight chunks.
% 中文翻译：上述发展说明向量级编码是有前景的，但同时也留下了一个问题：二值潜变量 codec 应该如何针对预训练 LLM 权重进行优化。Lookup-free binary quantization 可以从存储中移除显式码本，但无约束二值潜变量在训练过程中仍可能耦合码方向和潜变量幅值。BSQ 在视觉 tokenization 中通过先将潜变量归一化到超球面、再进行二值化来缓解这一问题，使离散码主要表示方向，同时让解码器接收尺度受控的二值输入。这个性质与权重压缩尤其相关：它不是把每个权重块映射到固定格点或显式中心，而是用球面二值码提供紧凑的无码本载荷，并由共享解码器学习一组权重块的重建规则。

However, simply increasing the length of a one-stage spherical binary code is not necessarily an efficient way to use the available bit budget. For a linear weight matrix $\mathbf{W}\in\mathbb{R}^{d_{\mathrm{out}}\times d_{\mathrm{in}}}$ partitioned into $d$-dimensional chunks, the number of training vectors for the codec is only $d_{\mathrm{out}}d_{\mathrm{in}}/d$, whereas the implicit binary code space contains $2^{b}$ possible sign patterns for a $b$-bit code. In typical Transformer projections, this gap can be large when $b$ is 32 or 64; many code patterns are not exercised by the fixed set of weight chunks. Therefore, a larger implicit code space does not by itself guarantee better effective reconstruction. To make the additional code capacity more effectively used, we design a second-stage residual BSQ codec that assigns extra bits to the residual structure not captured by the base codec, rather than simply enlarging a single-stage code.
% 中文翻译：然而，单纯增加一阶段球面二值码长度并不一定是高效利用比特预算的方式。对于一个线性权重矩阵 $\mathbf{W}\in\mathbb{R}^{d_{\mathrm{out}}\times d_{\mathrm{in}}}$，如果将其划分为 $d$ 维权重块，那么用于训练 codec 的向量数量只有 $d_{\mathrm{out}}d_{\mathrm{in}}/d$，而一个 $b$ bit 码对应的隐式二值码空间包含 $2^{b}$ 种符号模式。在典型 Transformer projection 中，当 $b$ 为 32 或 64 时，这一差距可能很大；固定权重块集合并不会使用许多码模式。因此，更大的隐式码空间本身并不保证更好的有效重建。为了更有效地利用额外的编码容量，我们设计了二阶残差 BSQ codec，将额外比特分配给基础 codec 未捕获的残差结构，而不是简单地扩大单阶段码。

Beyond the codec itself, extreme compression also requires a recovery objective that is aligned with the final model behavior. Layer-wise distillation is memory-efficient, but a local reconstruction or hidden-state loss for one layer is not necessarily aligned with perplexity or downstream accuracy after all compressed layers are assembled. The mismatch can be amplified when many independently optimized layers are composed. We therefore organize recovery at the Transformer-module-category level: after replacing one category of linear modules, the model is distilled with the compressed category active, so that the recovery loss reflects the interaction between that category and the remaining full model. This design keeps the memory cost lower than full end-to-end retraining while avoiding a purely layer-local recovery objective.
% 中文翻译：除了 codec 本身，极低比特压缩还需要一个与最终模型行为相匹配的精度恢复目标。逐层蒸馏具有显存效率优势，但单层的局部重建损失或隐藏状态损失，在所有压缩层组装之后并不一定与困惑度或下游任务精度一致。当许多独立优化的层被组合起来时，这种不匹配可能会被放大。因此，我们将恢复过程组织在 Transformer 模块类别层面：每替换一类 linear 模块后，就在该压缩类别生效的情况下进行蒸馏，使恢复损失能够反映该类别与剩余完整模型之间的交互。这个设计比完整端到端重训练具有更低显存成本，同时避免完全局部的逐层恢复目标。

Together, these components form a 2-bit-oriented LLM compression pipeline rather than a codec-only construction. On Qwen3-8B, the final compressed model achieves an average downstream accuracy within 2 percentage points of the FP16/BF16 model under the reported real storage budget. 
% \textcolor{blue}{[Verify the exact real bpw, benchmark suite, and final accuracy gap before submission.]}
% 中文翻译：这些组件共同形成了一个面向 2 bit 的 LLM 压缩流程，而不仅仅是一个单独的 codec 设计。在 Qwen3-8B 上，最终压缩模型在所报告的真实存储预算下，平均下游精度与 FP16/BF16 模型的差距控制在 2 个百分点以内。[投稿前需要核对准确的真实 bpw、评测任务集合和最终精度差距。]

The main contributions of this paper are summarized as follows:
% 中文翻译：本文的主要贡献总结如下：
\begin{itemize}
    \item We present \method{}, a storage-aware 2-bit-oriented compression pipeline for LLM weights. The pipeline integrates codebook-free binary spherical coding, residual code allocation, category-wise replacement, and recovery distillation, so that the compressed model can be optimized and evaluated under an explicit real-size budget.
    % 中文翻译：我们提出 BiSCo-LLM，一个面向 LLM 权重的存储感知 2-bit 压缩流程。该流程将无码本二值球面编码、残差码分配、逐类别替换和恢复蒸馏统一起来，使压缩模型可以在明确的真实尺寸预算下被优化和评估。
    \item We introduce a second-stage residual BSQ codec to improve the use of additional code capacity. Rather than simply enlarging a one-stage binary code, \method{} assigns extra bits to the residual structure not captured by the base spherical codec, yielding a practical rate--distortion path under explicit storage accounting.
    % 中文翻译：我们提出二阶残差 BSQ codec，以更有效地利用额外编码容量。BiSCo-LLM 不是简单扩大一阶段二值码，而是将额外比特分配给基础球面 codec 未捕获的残差结构，从而在显式存储核算下形成实际可用的率失真路径。
    \item We design category-wise codec optimization and recovery distillation for Transformer linear modules. The codec is trained and applied by module category, and recovery is performed after each category replacement, so that the distillation objective better reflects the interaction between the compressed category and the remaining model than purely layer-wise recovery.
    % 中文翻译：我们为 Transformer 线性模块设计逐类别 codec 优化与恢复蒸馏。codec 按模块类别训练和应用，并在每个类别替换后执行恢复，使蒸馏目标相比纯逐层恢复更能反映压缩类别与剩余模型之间的交互。
    \item We evaluate the proposed pipeline under real storage budgets, explicitly accounting for BSQ code streams, decoder parameters, metadata, optional protected-channel payloads, LoRA adapters, and decoding cost. On Qwen3-8B, the final compressed model is reported to retain average downstream accuracy within 2 percentage points of the FP16/BF16 model. 
    % \textcolor{blue}{[Verify exact bpw and benchmark numbers before submission.]}
    % 中文翻译：我们在真实存储预算下评估所提出的流程，显式核算 BSQ 码流、解码器参数、元数据、可选重要通道载荷、LoRA adapter 和解码开销。在 Qwen3-8B 上，最终压缩模型被报告为平均下游精度与 FP16/BF16 模型相差不超过 2 个百分点。[投稿前需要核对准确 bpw 和 benchmark 数值。]
\end{itemize}

The rest of the paper is organized as follows. Section~\ref{sec:related} reviews LLM quantization, VQ-based LLM compression, codebook-free discrete representations, and sensitivity-aware compression. Section~\ref{sec:method} presents the proposed framework. Section~\ref{sec:experiments} describes the experimental protocol, baseline design, and ablation plan. Section~\ref{sec:discussion} discusses limitations and implementation choices. Section~\ref{sec:conclusion} concludes the paper.
% 中文翻译：本文剩余部分组织如下。第二节回顾 LLM 量化、基于 VQ 的 LLM 压缩、无码本离散表示和敏感性感知压缩。第三节介绍所提出的框架。第四节描述实验协议、基线设计和消融计划。第五节讨论局限性和实现选择。第六节总结全文。

\section{Related Work}\label{sec:related}
\subsection{Low-Bit Quantization of Large Language Models}

LLM quantization reduces storage, memory bandwidth, and serving cost by replacing high-precision weights, activations, or cached states with low-precision representations. Early LLM-specific studies mainly focused on post-training quantization (PTQ), which avoids full retraining and is therefore attractive for compressing large pretrained models. ZeroQuant combines fine-grained quantization, layer-wise distillation, and optimized inference kernels for large Transformers \cite{yao2022zeroquant}. LLM.int8() identifies emergent outlier features as a major obstacle for scaling 8-bit matrix multiplication to large Transformer models \cite{dettmers2022llmint8}. GPTQ introduces an approximate second-order, layer-wise weight-only PTQ procedure and has become a standard baseline for low-bit LLM compression \cite{frantar2022gptq}. RPTQ further reorders channels to reduce activation range imbalance \cite{yuan2023rptq}. SmoothQuant migrates activation difficulty into weights through equivalent scaling, enabling accurate weight--activation quantization \cite{xiao2022smoothquant}. AWQ protects activation-salient weights through activation-aware scaling, while OmniQuant learns quantization-friendly equivalent transformations under calibration data \cite{lin2023awq,shao2023omniquant}.
% 中文翻译：LLM 量化通过用低精度表示替代高精度权重、激活或缓存状态，降低存储、显存带宽和服务成本。早期面向 LLM 的研究主要关注后训练量化，即避免完整重训练，因此适合压缩大规模预训练模型。ZeroQuant 结合细粒度量化、逐层蒸馏和优化推理 kernel 来处理大规模 Transformer。LLM.int8() 指出 emergent outlier features 是将 8-bit 矩阵乘法扩展到大规模 Transformer 时的主要障碍。GPTQ 引入近似二阶的逐层 weight-only PTQ 流程，并成为低比特 LLM 压缩中的标准基线。RPTQ 进一步通过通道重排降低激活范围不均衡。SmoothQuant 通过等价缩放将激活端的量化困难迁移到权重端，从而实现准确的权重和激活联合量化。AWQ 通过 activation-aware scaling 保护激活显著权重，而 OmniQuant 在校准数据上学习有利于量化的等价变换。

More recent work shows that low-bit LLM quantization is not a single design choice, but a combination of optimization objective, outlier handling, transformation design, and training budget. OWQ stores a small subset of sensitive weak columns in high precision and quantizes the remaining dense weights, showing the benefit of structured outlier-aware mixed precision \cite{lee2024owq}. SpQR similarly combines a sparse high-precision component with a quantized dense component for near-lossless weight compression \cite{dettmers2023spqr}. SqueezeLLM uses sensitivity-based non-uniform quantization together with dense-and-sparse decomposition \cite{kim2023squeezellm}. Atom studies low-bit serving with mixed precision and fine-grained quantization for efficient deployment \cite{zhao2023atom}. Recent transformation-based methods further reduce outlier difficulty before quantization: QuaRot and SpinQuant use equivalent rotations to make weights and activations easier to quantize, while AffineQuant extends the transformation space from scaling or rotation to learnable affine transformations \cite{ashkboos2024quarot,liu2024spinquant,ma2024affinequant}.
% 中文翻译：近年的工作表明，低比特 LLM 量化并不是单一设计选择，而是优化目标、离群值处理、变换设计和训练预算的组合。OWQ 将少量敏感 weak columns 以高精度保存，并对其余密集权重进行量化，体现了结构化离群感知混合精度的作用。SpQR 也将稀疏高精度部分与量化密集部分结合，用于近似无损的权重压缩。SqueezeLLM 使用基于敏感性的非均匀量化以及 dense-and-sparse 分解。Atom 面向高效部署，研究混合精度和细粒度量化下的低比特服务。近期的变换类方法则进一步在量化前降低离群值带来的困难：QuaRot 和 SpinQuant 使用等价旋转使权重和激活更易量化，而 AffineQuant 将变换空间从缩放或旋转扩展到可学习的仿射变换。

Another line of work uses quantization-aware training(QAT) or distillation to recover accuracy at more aggressive bit widths. LLM-QAT investigates data-free quantization-aware training for LLMs and also quantizes the KV cache, showing that training-based recovery becomes important when PTQ breaks down at lower precision \cite{liu2024llmqat}. BitDistiller combines quantization-aware training with self-distillation to improve sub-4-bit LLMs \cite{du2024bitdistiller}. OneBit pushes weight compression toward approximately 1-bit representation by introducing a sign-value-independent decomposition and quantization-aware knowledge distillation \cite{xu2024onebit}. These methods are closely related to the extreme-compression regime targeted by \method{}, but they rely on scalar or specially structured parameter representations. In contrast, \method{} uses binary spherical codes with neural residual decoders. It is therefore closer to learned discrete compression than to conventional scalar quantization, while still inheriting the central lessons from LLM quantization: outliers must be treated carefully, calibration statistics are useful, and the storage overhead of any auxiliary component must be counted explicitly.
% 中文翻译：另一类工作使用量化感知训练或蒸馏来恢复更激进比特宽度下的精度。LLM-QAT 研究面向 LLM 的 data-free quantization-aware training，并同时量化 KV cache，说明当 PTQ 在更低精度下失效时，训练式恢复变得重要。BitDistiller 将量化感知训练与自蒸馏结合，用于提升 sub-4-bit LLM 的性能。OneBit 通过 sign-value-independent decomposition 和量化感知知识蒸馏，将权重压缩推进到近似 1-bit 表示。这些方法与 BiSCo-LLM 关注的极限压缩场景密切相关，但它们依赖标量量化或特殊结构化参数表示。相比之下，BiSCo-LLM 使用二值球面码和神经残差解码器，因此更接近学习式离散压缩，而不是传统标量量化；但它仍然继承了 LLM 量化中的核心经验：必须谨慎处理离群值，校准统计是有用的，并且任何辅助组件的存储开销都必须被显式计入。

\subsection{Vector-Quantized and Structured-Code Weight Compression}

Vector quantization (VQ) is a classical compression paradigm in which a vector is represented by the index of a codeword rather than by independently quantized scalar entries. It has also become a standard building block of neural discrete representation learning, as exemplified by VQ-VAE \cite{oord2017vqvae}. For LLM weight compression, VQ is attractive because a local weight block can be encoded as one or several discrete symbols, thereby increasing the effective representational dimension under a fixed bit budget. This property is especially important below 3 bits per weight, where scalar quantizers have only a few reconstruction levels and can no longer flexibly match the anisotropic distribution of pretrained Transformer weights.
% 中文翻译：向量量化是一种经典压缩范式，其中一个向量由码字索引表示，而不是由独立量化的标量条目表示。它也是神经离散表示学习中的标准组件，例如 VQ-VAE。对于 LLM 权重压缩，VQ 很有吸引力，因为局部权重块可以被编码为一个或多个离散符号，从而在固定比特预算下提高有效表示维度。这一点在低于 3 bit/weight 时尤其重要，因为标量量化器只有很少的重建取值，已经难以灵活匹配预训练 Transformer 权重的各向异性分布。

Recent LLM compression methods have explored several forms of vector or structured-code quantization. AQLM revisits extreme LLM compression from the perspective of multi-codebook quantization and represents weights through learned additive codebooks, with joint optimization across Transformer blocks \cite{egiazarian2024aqlm}. GPTVQ further studies post-training VQ at higher vector dimensions and interleaves vector quantization with Hessian-informed updates to the remaining unquantized weights \cite{vanbaalen2024gptvq}. These methods show that increasing the quantization dimension can substantially improve the size--accuracy trade-off compared with purely scalar rounding, but their compressed representation still depends on explicit codebooks and index lookup.
% 中文翻译：近期 LLM 压缩方法探索了多种向量量化或结构化编码形式。AQLM 从多码本量化的角度重新审视极限 LLM 压缩，通过学习式加性码本表示权重，并在 Transformer block 内联合优化。GPTVQ 进一步研究更高向量维度下的后训练 VQ，并将向量量化与 Hessian 信息引导的未量化权重更新交替进行。这些方法表明，相比纯标量舍入，提高量化维度可以显著改善模型大小和精度之间的折中，但其压缩表示仍然依赖显式码本和索引查表。

Another important direction is to design structured codebooks or coding rules with better theoretical or hardware properties. QuIP introduces incoherence processing so that weight and Hessian directions become less aligned with coordinate axes before quantization, enabling viable 2-bit LLM quantization with theoretical guarantees\cite{chee2023quip}. QuIP\# improves this line by using randomized Hadamard transforms and lattice codebooks, especially the highly symmetric $E_8$ lattice, for more effective vector quantization in extreme compression regimes\cite{tseng2024quipsharp}. QTIP replaces explicit low-dimensional VQ lookup with trellis-coded quantization, using a stateful decoder to decouple the effective quantization dimension from the explicit codebook size \cite{tseng2024qtip}. These works indicate that the coding structure, not only the nominal bit-width, is critical for low-bit LLM quality.
% 中文翻译：另一个重要方向是设计具有更好理论性质或硬件性质的结构化码本或编码规则。QuIP 引入 incoherence processing，使权重和 Hessian 方向在量化前与坐标轴更不对齐，从而实现具有理论保证的可用 2-bit LLM 量化。QuIP# 在这一方向上进一步使用随机 Hadamard 变换和格点码本，尤其是高度对称的 $E_8$ lattice，用于极限压缩场景中的更有效向量量化。QTIP 用 trellis-coded quantization 替代显式低维 VQ 查表，通过有状态解码器将有效量化维度与显式码本大小解耦。这些工作说明，对于低比特 LLM 质量而言，编码结构本身与名义比特宽度同样关键。

More recent work also incorporates second-order objectives, residual quantization, and outlier handling into vector PTQ. VPTQ formulates LLM vector quantization with second-order optimization, refines weights with channel-independent second-order updates, and extends the framework to residual and outlier quantization \cite{liu2024vptq}. This is closely related to our use of multi-stage residual coding and important-channel protection. However, VPTQ still follows the conventional VQ view in which vectors are compressed into indices associated with lookup tables. In contrast, \method{} stores bit-packed binary spherical codes and reconstructs weights through compact neural decoders, so the main reconstruction capacity is carried by shared decoder parameters rather than by explicit vector codebooks.
% 中文翻译：近期工作也将二阶目标、残差量化和离群值处理引入向量 PTQ。VPTQ 用二阶优化表述 LLM 向量量化问题，通过 channel-independent second-order updates 细化权重，并将框架扩展到残差和离群值量化。这与我们使用多级残差编码和重要通道保护密切相关。然而，VPTQ 仍然遵循传统 VQ 视角，即向量被压缩为与查找表关联的索引。相比之下，BiSCo-LLM 存储打包后的二值球面码，并通过紧凑神经解码器重建权重，因此主要重建能力由共享解码器参数承担，而不是由显式向量码本承担。

LiftQuant is another close concurrent work. It uses dimensional lifting and projection, where a low-dimensional weight vector is approximated by projecting a simple 1-bit lattice from a higher-dimensional lifted space \cite{he2026liftquant}. Since the effective bit-width is controlled by the ratio between the lifted and original dimensions, LiftQuant provides quasi-continuous bit-width control and can better fit specific memory budgets. Compared with LiftQuant, \method{} does not rely on a fixed linear lift-and-project reconstruction path. Instead, it learns nonlinear BSQ decoders for Transformer module categories, adds residual stages to improve the rate--distortion curve, and assigns additional compressed residual capacity to important channels. Therefore, our method can be viewed as a storage-aware neural coding approach rather than a fixed structured-code projection method.
% 中文翻译：LiftQuant 是另一个非常接近的同期工作。它使用维度提升和投影，即通过从高维提升空间中投影简单的 1-bit lattice 来近似低维权重向量。由于有效比特宽度由提升维度和原始维度的比例控制，LiftQuant 提供了准连续的比特宽度控制，并能更好适配特定显存预算。与 LiftQuant 相比，BiSCo-LLM 不依赖固定的线性 lift-and-project 重建路径。相反，它为 Transformer 模块类别学习非线性 BSQ 解码器，加入残差阶段改善率失真曲线，并为重要通道分配额外的压缩残差容量。因此，我们的方法可以被视为一种存储感知的神经编码方法，而不是固定结构化编码投影方法。

UniSVQ and LC-QAT are also closely related because they reformulate low-bit LLM compression through structured vector quantization. UniSVQ parameterizes vector codewords as affine transformations of integer-lattice points, thereby providing a unified view of scalar and vector quantization \cite{wang2026unisvq}. LC-QAT studies data-efficient 2-bit quantization-aware training with linear-constrained vector quantization, using a constrained code structure and training data to recover accuracy \cite{wang2026lcqat}. In contrast, \method{} does not store or optimize an explicit lattice codebook as the primary representation. It stores bit-packed binary spherical codes and uses compact category-wise neural decoders to reconstruct weights. Moreover, the second-stage residual BSQ codec and category-wise recovery distillation separate representation capacity from model-behavior recovery, which differs from methods that mainly rely on a single structured vector quantizer or QAT objective.
% 中文翻译：UniSVQ 和 LC-QAT 也与本文密切相关，因为它们都通过结构化向量量化重新表述低比特 LLM 压缩。UniSVQ 将向量码字参数化为整数格点的仿射变换，从而提供了连接标量量化和向量量化的统一视角。LC-QAT 研究带线性约束向量量化的数据高效 2-bit 量化感知训练，使用受约束的编码结构和训练数据恢复精度。相比之下，BiSCo-LLM 不以存储或优化显式格码本作为主要表示。它存储打包后的二值球面码，并使用紧凑的逐类别神经 decoder 重建权重。此外，第二阶段残差 BSQ codec 和逐类别恢复蒸馏将表示容量与模型行为恢复分开处理，这不同于主要依赖单一结构化向量量化器或 QAT 目标的方法。

Overall, vector-quantized and structured-code compression methods demonstrate that extreme low-bit LLM compression requires more expressive local representations than scalar quantization. Nevertheless, explicit codebooks, lattice lookup rules, trellis decoders, or fixed projection matrices introduce their own storage, implementation, or adaptation constraints. \method{} follows the same high-dimensional coding motivation, but removes explicit codebooks and uses amortized neural decoders shared across module categories. This design is intended to preserve the representational advantage of vector coding while making the storage accounting of binary codes, residual stages, decoders, and compensation modules explicit.
% 中文翻译：总体而言，向量量化和结构化编码压缩方法表明，极低比特 LLM 压缩需要比标量量化更有表达力的局部表示。然而，显式码本、格点查表规则、trellis 解码器或固定投影矩阵都会引入各自的存储、实现或适配约束。BiSCo-LLM 遵循相同的高维编码动机，但去除了显式码本，并使用在模块类别间共享的摊销神经解码器。该设计旨在保留向量编码的表示优势，同时显式统计二值码、残差阶段、解码器和补偿模块的存储开销。
\subsection{Codebook-Free Discrete Representation Learning}

Classical neural discrete representation learning usually relies on an explicit codebook. VQ-VAE maps each encoder output to the nearest learned embedding vector and trains the encoder--decoder system with straight-through gradient estimation and auxiliary commitment losses \cite{oord2017vqvae}. This design has been highly influential in image, audio, and video tokenization, but it also introduces well-known practical issues: the codebook may be under-utilized, some entries may collapse or rarely be selected, nearest-neighbor lookup becomes expensive when the vocabulary is large, and the codebook itself becomes an additional storage component. These issues are acceptable in many data tokenization settings, but they become more restrictive when discrete representation learning is used as a compression mechanism for pretrained model weights, where every auxiliary parameter must be included in the final model size.
% 中文翻译：经典神经离散表示学习通常依赖显式码本。VQ-VAE 将每个编码器输出映射到最近的可学习 embedding 向量，并通过 straight-through 梯度估计和额外的 commitment loss 训练编码器-解码器系统。该设计在图像、音频和视频 tokenization 中影响很大，但也带来了若干已知实践问题：码本可能利用不足，一些条目可能塌缩或很少被选中，当词表很大时最近邻查找开销较高，并且码本本身也会成为额外存储组件。这些问题在许多数据 tokenization 场景中是可以接受的，但当离散表示学习被用于预训练模型权重压缩时会更加受限，因为所有辅助参数都必须计入最终模型大小。

A recent response to these limitations is to replace learned vector codebooks with simpler implicit or non-parametric discrete spaces. FSQ replaces vector quantization by finite scalar levels: the latent is projected to a low-dimensional vector, and each dimension is quantized to a small fixed set of values, yielding an implicit Cartesian-product codebook without learned entries \cite{mentzer2023fsq}. MAGVIT-v2 introduces lookup-free quantization (LFQ), where visual tokens are represented by binary latent codes rather than by nearest-neighbor lookup in a learned embedding table \cite{yu2024magvitv2}. This makes it possible to use very large effective vocabularies while avoiding explicit vector-codebook search. Open-MAGVIT2 further demonstrates the practical value of this direction by providing an open-source replication and scaling study of MAGVIT-v2-style tokenizers \cite{luo2024openmagvit2}.
% 中文翻译：针对这些限制，近期一个重要方向是用更简单的隐式或非参数离散空间替代可学习向量码本。FSQ 用有限标量等级替代向量量化：先将 latent 投影到低维向量，再将每个维度量化到一组固定取值，从而得到一个没有可学习条目的隐式笛卡尔积码本。MAGVIT-v2 引入 lookup-free quantization，即视觉 token 由二值潜变量表示，而不是通过在可学习 embedding table 中进行最近邻查找得到。这使得模型可以使用非常大的有效词表，同时避免显式向量码本搜索。Open-MAGVIT2 进一步通过开源复现和扩展研究展示了 MAGVIT-v2 式 tokenizer 的实践价值。

Binary tokenizers push this idea further by treating the bit representation itself as the discrete latent object. BSQ projects the latent feature to a hypersphere and then applies binary quantization, producing compact binary spherical codes without an explicit codebook \cite{zhao2024bsq}. MaskBit shows that bit tokens can also be used directly by the generative model, removing the need for an additional token embedding table in the second-stage generator \cite{weber2024maskbit}. These works are especially relevant to \method{} because they suggest that binary codes are not merely implementation-level bit strings, but can serve as semantically meaningful and scalable discrete representations.
% 中文翻译：二值 tokenizer 进一步推进了这一思想，即直接将 bit 表示本身视为离散潜变量。BSQ 将 latent feature 投影到超球面上，然后进行二值量化，从而在没有显式码本的情况下得到紧凑的二值球面码。MaskBit 表明 bit token 也可以被生成模型直接使用，从而去除第二阶段生成器中的额外 token embedding table。这些工作与 BiSCo-LLM 尤其相关，因为它们说明二值码并不只是实现层面的 bit string，也可以作为具有语义和可扩展性的离散表示。

There are also complementary attempts to stabilize or restructure VQ training rather than remove the codebook completely. SimVQ reparameterizes code vectors through a learnable linear transformation to alleviate representation collapse \cite{zhu2024simvq}. The rotation trick modifies the gradient path through VQ by encoding the relative angle and magnitude between encoder outputs and selected code vectors into the backward signal \cite{fifty2024rotation}. Grouped Spherical Quantization (GSQ) studies spherical codebook initialization and grouped latent scaling for image tokenizers \cite{wang2024gsq}. These methods indicate that the geometry of the discrete latent space, the gradient estimator, and the utilization pattern of codes are all important for high-quality discrete representation learning.
% 中文翻译：也有一些互补工作并不完全去除码本，而是试图稳定或重构 VQ 训练。SimVQ 通过可学习线性变换重新参数化码向量，以缓解表示塌缩。rotation trick 通过在反向传播信号中编码 encoder output 和所选 code vector 之间的相对角度与模长，改变 VQ 层的梯度路径。Grouped Spherical Quantization 研究用于图像 tokenizer 的球面码本初始化和分组 latent 扩展。这些方法说明，离散潜空间的几何结构、梯度估计方式和码的利用模式都会影响高质量离散表示学习。

Despite these connections, applying codebook-free discrete representation learning to LLM weight compression is fundamentally different from image or video tokenization. Visual tokenizers learn an encoder and a decoder for a distribution of input samples, and the learned tokens are consumed by a downstream generative model. In contrast, LLM weight compression encodes a fixed set of pretrained tensors, and the decoder is part of the deployed compressed model. Therefore, decoder parameters, residual stages, per-category sharing, and any compensation modules must be counted in the storage budget. Moreover, the reconstruction target is not perceptual fidelity but preservation of the pretrained model function, so small weight-space errors can cause large downstream accuracy degradation.
% 中文翻译：尽管这些工作与本文有关，但将无码本离散表示学习用于 LLM 权重压缩与图像或视频 tokenization 有本质区别。视觉 tokenizer 为输入样本分布学习编码器和解码器，得到的 token 会被下游生成模型使用。相比之下，LLM 权重压缩编码的是一组固定的预训练张量，解码器本身就是部署后压缩模型的一部分。因此，解码器参数、残差阶段、按类别共享策略以及任何补偿模块都必须计入存储预算。此外，重建目标不是感知保真度，而是保持预训练模型功能，因此权重空间中的小误差也可能导致下游精度显著下降。

\method{} adapts the codebook-free idea to this weight-compression setting. Instead of storing learned codewords or performing nearest-neighbor lookup, it stores bit-packed binary spherical codes and reconstructs weight blocks through compact neural decoders. To make this feasible for LLMs, \method{} further introduces multi-stage residual coding, module-category-level codec organization, an auxiliary 8-bit path for sensitive channels, and optional LoRA compensation. In this sense, our method inherits the parameter-efficient and scalable spirit of FSQ, LFQ, and BSQ, but shifts the application target from data tokenization to storage-aware compression of pretrained LLM weights.
% 中文翻译：BiSCo-LLM 将无码本思想适配到权重压缩场景。它不存储可学习码字，也不执行最近邻查找，而是存储打包后的二值球面码，并通过紧凑神经解码器重建权重块。为了使其适用于 LLM，BiSCo-LLM 进一步引入多级残差编码、按模块类别组织的 codec、面向敏感通道的辅助 8bit 路径以及可选 LoRA 补偿。从这个意义上说，我们的方法继承了 FSQ、LFQ 和 BSQ 的参数高效性与可扩展性，但将应用目标从数据 tokenization 转向预训练 LLM 权重的存储感知压缩。

\subsection{Outliers, Saliency, and Low-Rank Compensation}

Outliers and saliency imbalance are central obstacles in LLM compression. LLM.int8() observes that large Transformer models exhibit emergent outlier features, and handles these dimensions separately to enable stable 8-bit matrix multiplication \cite{dettmers2022llmint8}. SmoothQuant further shows that activation outliers can be partially migrated into weights by equivalent scaling transformations, thereby making weight--activation quantization more tractable \cite{xiao2022smoothquant}. AWQ identifies activation-salient weights and protects them through activation-aware scaling, while OWQ keeps a small number of weak columns in higher precision to reduce the damage caused by outlier-associated sensitive channels \cite{lin2023awq,lee2024owq}. These studies indicate that the main difficulty in low-bit LLM compression is not merely the average quantization error, but the highly non-uniform distribution of error sensitivity across channels, layers, and module types.
% 中文翻译：离群值和显著性不均衡是 LLM 压缩中的核心障碍。LLM.int8() 观察到大规模 Transformer 模型中会出现 emergent outlier features，并对这些维度进行单独处理，从而实现稳定的 8-bit 矩阵乘法。SmoothQuant 进一步表明，可以通过等价缩放变换将激活离群值部分迁移到权重中，使权重和激活联合量化更易处理。AWQ 识别 activation-salient weights，并通过 activation-aware scaling 对其进行保护；OWQ 则将少量 weak columns 以更高精度保存，以降低与离群激活相关的敏感通道造成的损害。这些研究说明，低比特 LLM 压缩的主要困难不只是平均量化误差，而是误差敏感性在通道、层和模块类型之间高度不均匀。

A closely related line of work studies saliency for pruning and mixed compression. SparseGPT formulates one-shot LLM pruning as a layer-wise reconstruction problem with approximate second-order updates, and also shows that sparsity can be combined with weight quantization \cite{frantar2023sparsegpt}. Wanda proposes a simpler saliency metric based on the product of weight magnitude and input activation norm, requiring neither retraining nor explicit weight update \cite{sun2024wanda}. LLM-Pruner performs structural pruning by identifying non-critical coupled structures using gradient information and then recovers performance with lightweight tuning \cite{ma2023llmpruner}. Although these methods target pruning rather than dense low-bit coding, they provide an important lesson for extreme compression: channels or structures should not be treated uniformly, and calibration activations or gradients can provide useful proxies for preserving model function.
% 中文翻译：一个密切相关的方向是研究用于剪枝和混合压缩的显著性。SparseGPT 将一次性 LLM 剪枝表述为带近似二阶更新的逐层重建问题，并表明稀疏性可以与权重量化结合。Wanda 提出了一种更简单的显著性指标，即权重幅值与输入激活范数的乘积，且不需要重训练或显式权重更新。LLM-Pruner 使用梯度信息识别非关键的耦合结构进行结构化剪枝，并通过轻量调优恢复性能。尽管这些方法主要面向剪枝而不是密集低比特编码，但它们为极限压缩提供了重要经验：不应将所有通道或结构均匀处理，校准激活或梯度可以作为保持模型功能的有效代理信号。

Second-order and activation-aware criteria have also been widely adopted in quantization. GPTQ uses an approximate Hessian-based layer-wise objective to reduce the loss increase induced by weight quantization \cite{frantar2022gptq}. OmniQuant learns equivalent transformations under calibration data to make both weights and activations more quantization-friendly \cite{shao2023omniquant}. VPTQ extends vector post-training quantization with second-order optimization, residual quantization, and outlier handling \cite{liu2024vptq}. In \method{}, these observations are used more conservatively: the BSQ codec itself is trained with MSE reconstruction, while calibration statistics are used only to select a small optional 8-bit protected set.
% 中文翻译：二阶准则和激活感知准则也被广泛用于量化。GPTQ 使用近似 Hessian 的逐层目标来降低权重量化引起的 loss 增加。OmniQuant 在校准数据上学习等价变换，使权重和激活都更加量化友好。VPTQ 将向量后训练量化扩展到二阶优化、残差量化和离群值处理。在 BiSCo-LLM 中，我们更保守地使用这些观察：BSQ codec 本身仍采用 MSE 重建训练，而校准统计仅用于选择少量可选 8bit 保护通道。

Low-rank adaptation provides another way to compensate for compression error. LoRA inserts trainable low-rank adapters into pretrained models and has become a standard parameter-efficient fine-tuning technique \cite{hu2021lora}. QLoRA shows that low-rank tuning can be combined with a quantized base model by using 4-bit quantization and memory-efficient optimization \cite{dettmers2023qlora}. More recent methods more explicitly couple low-rank adaptation with quantization. QA-LoRA performs quantization-aware low-rank adaptation by balancing the degrees of freedom between quantization and adaptation through group-wise operators \cite{xu2024qalora}. LoftQ jointly quantizes the pretrained model and initializes LoRA adapters to reduce the discrepancy between quantized and full-precision models \cite{li2024loftq}. LQ-LoRA decomposes each weight matrix into a fixed quantized component and a trainable high-precision low-rank component under memory constraints \cite{guo2023lqlora}. L4Q further integrates quantization-aware training with LoRA and reduces QAT memory overhead to a level comparable to LoRA fine-tuning \cite{jeon2025l4q}.
% 中文翻译：低秩适配提供了另一种补偿压缩误差的方式。LoRA 在预训练模型中插入可训练低秩适配器，并已成为标准的参数高效微调技术。QLoRA 表明，可以通过 4-bit 量化和显存高效优化，将低秩微调与量化基座模型结合。近期方法更明确地耦合低秩适配与量化。QA-LoRA 通过 group-wise operators 平衡量化和适配的自由度，实现量化感知低秩适配。LoftQ 同时量化预训练模型并初始化 LoRA adapter，以减小量化模型与全精度模型之间的差距。LQ-LoRA 在显存约束下将每个权重矩阵分解为固定量化部分和可训练高精度低秩部分。L4Q 进一步将量化感知训练与 LoRA 结合，并将 QAT 的显存开销降低到接近 LoRA 微调的水平。

There is also growing interest in using low-rank modules specifically as quantization-error compensators rather than as ordinary task adapters. RILQ studies LoRA-based quantization error compensation in the 2-bit regime and points out that conventional low-rank compensation can become ineffective when the quantization error is too aggressive \cite{lee2025rilq}. SERQ further explores saliency-aware low-rank error reconstruction for low-bit LLM inference \cite{park2026serq}. These works are particularly relevant to \method{} because our LoRA modules are not intended to define the primary compressed representation. Instead, the main representation is the BSQ code and neural decoder, while LoRA is used as an optional compensation path after codebook-free reconstruction.
% 中文翻译：也有越来越多工作将低秩模块专门用作量化误差补偿器，而不是普通任务适配器。RILQ 研究 2-bit 场景下的 LoRA-based quantization error compensation，并指出当量化误差过于激进时，传统低秩补偿可能变得无效。SERQ 进一步研究面向低比特 LLM 推理的 saliency-aware low-rank error reconstruction。这些工作与 BiSCo-LLM 尤其相关，因为我们的 LoRA 模块并不用于定义主要压缩表示。相反，主要表示由 BSQ code 和神经解码器承担，而 LoRA 只是无码本重建之后的可选补偿路径。

In summary, existing work suggests three design principles that are directly reflected in \method{}: salient channels may need stronger protection, calibration statistics can help identify these channels, and low-rank modules should be reported separately from the base compressed representation. Accordingly, \method{} uses MSE-based BSQ codec training, an auxiliary 8-bit path for a small set of sensitive channels, and separate reporting of codec-only, category-level LoRA, and end-to-end LoRA results. This separation is important for fair evaluation, since otherwise the effect of the proposed binary spherical codec would be conflated with the effect of auxiliary high-precision storage or trainable compensation parameters.
% 中文翻译：总结来看，已有工作给出了三个直接体现在 BiSCo-LLM 中的设计原则：显著通道可能需要更强保护，校准统计可以帮助识别这些通道，低秩模块应与基础压缩表示分开报告。因此，BiSCo-LLM 使用基于 MSE 的 BSQ codec 训练、面向少量敏感通道的辅助 8bit 路径，并分别报告仅编解码器、类别级 LoRA 和端到端 LoRA 的结果。这种区分对于公平评估很重要，否则所提出二值球面编解码器的作用会与辅助高精度存储或可训练补偿参数的作用混在一起。

\section{Method}\label{sec:method}
\subsection{Problem Formulation and Bit Accounting}
Let $\mathbf{W}_{\ell}^{c}\in\R^{m_{\ell}^{c}\times n_{\ell}^{c}}$ denote the weight matrix of layer $\ell$ and module category $c$. We partition this matrix into chunks $\mathbf{x}_{\ell,i}^{c}\in\R^{d}$, where $d$ is the chunk dimension. The goal is to store each chunk with a binary code $\mathbf{b}_{\ell,i}^{c}\in\{-1,+1\}^{b}$ and reconstruct it with a decoder:
% 中文翻译：令 W 表示第 ell 层、类别 c 的权重矩阵。我们将这个矩阵划分为 d 维权重块，其中 d 是权重块维度。目标是用一个 b 位二值码存储每个权重块，并使用解码器重建它：
\begin{equation}
    \widehat{\mathbf{x}}_{\ell,i}^{c}=D_{\theta_c}(\mathbf{b}_{\ell,i}^{c}).
\end{equation}

Ignoring decoder and metadata overhead, one BSQ stage has a nominal bit rate
% 中文翻译：如果暂时忽略解码器和元数据开销，一个 BSQ 阶段的名义比特率为：
\begin{equation}
    \mathrm{bpw}_{\mathrm{stage}}=\frac{b}{d}.
\end{equation}
Here $b$ corresponds to \texttt{codebook\_bits} and $d$ corresponds to \texttt{codebook\_dim}. For example, $b=32$ and $d=32$ gives one nominal bit per weight for one stage. With $S$ residual stages, the code-only rate is approximately $S b/d$ before adding decoder parameters, optional 8-bit protected channels, LoRA adapters, and metadata.
% 中文翻译：这里 b 对应 codebook_bits，d 对应 codebook_dim。例如，b=32 且 d=32 时，一个阶段的名义码率是每个权重 1 bit。如果有 S 个残差阶段，在加入解码器、残差保护、LoRA 和元数据之前，仅码流部分的码率大约是 S b/d。

The real compressed size is reported by a compact additive form:
% 中文翻译：真实压缩大小用一个紧凑的加性形式来报告：
\begin{equation}
    S_{\mathrm{tot}} = S_{\mathrm{code}} + S_{\mathrm{dec}} + S_{\mathrm{8bit}} + S_{\mathrm{LoRA}} + S_{\mathrm{meta}} .
    \label{eq:real_size}
\end{equation}
Here $S_{\mathrm{code}}$ denotes all bit-packed BSQ code streams, including the base stage and the second residual stage. $S_{\mathrm{dec}}$ denotes all neural decoder parameters, $S_{\mathrm{8bit}}$ denotes the optional 8-bit protected-channel payload, $S_{\mathrm{LoRA}}$ denotes optional LoRA adapters, and $S_{\mathrm{meta}}$ denotes indices, scales, and other metadata.
% 中文翻译：其中，S_code 表示所有打包后的 BSQ 码流，包括基础阶段和二阶残差阶段。S_dec 表示所有神经解码器参数，S_8bit 表示可选 8bit 保护通道载荷，S_LoRA 表示可选 LoRA 适配器，S_meta 表示索引、缩放因子和其他元数据。
This accounting is necessary because a method can have the same nominal bit rate but very different real model sizes. It is also the fair way to compare \method{} with explicit VQ methods and LiftQuant-style projection methods, where codebooks, projection parameters, scales, and metadata must be counted.
% 中文翻译：这种核算是必要的，因为不同方法即使具有相同的名义比特率，也可能有非常不同的真实模型大小。这也是将 BiSCo-LLM 与显式 VQ 方法和 LiftQuant 式投影方法公平比较的方式，因为码本、投影参数、缩放因子和元数据都必须被计入。

\subsection{Pipeline Overview}
Under this formulation, \method{} constructs a compressed model through a storage-aware category-wise pipeline. For each target module category, the corresponding weight chunks are first encoded by a base BSQ codec, which maps them to bit-packed unit-sphere sign codes and reconstructs them with a compact neural decoder. A second BSQ codec is then trained on the residual left by the base reconstruction, assigning additional code capacity to the weight structure that is not captured in the first stage. After a category has been replaced by its compressed modules, category-wise recovery distillation is performed with the compressed category active in the model, before the pipeline proceeds to the next category. Once all target categories have been compressed and locally recovered, a final end-to-end recovery distillation stage can be applied to the fully compressed model to further reduce accumulated cross-category mismatch. A lightweight 8-bit protected path can be optionally enabled for a very small set of sensitive channels, and its storage is counted separately from the main BSQ payload.
% 中文翻译：在上述定义下，BiSCo-LLM 通过一个存储感知的逐类别流程构造压缩模型。对于每个目标模块类别，首先使用基础 BSQ codec 对对应权重块进行编码，将其映射为打包后的单位球面符号码，并由紧凑神经解码器重建。随后，在基础重建留下的残差上训练第二个 BSQ codec，将额外编码容量分配给第一阶段未捕获的权重结构。某一类别被压缩模块替换后，在该压缩类别生效的模型中执行逐类别恢复蒸馏，然后再进入下一类别。当所有目标类别都完成压缩和局部恢复后，可以在完整压缩模型上进一步执行最终端到端恢复蒸馏，以降低跨类别累计错配。对于极少量敏感通道，可以可选启用轻量 8bit 保护路径，其存储开销与主 BSQ 载荷分开统计。

% \begin{figure*}[t]
% \centering
% \includegraphics[width=0.98\textwidth]{figures/bisco_pipeline.png}
% \caption{Storage-aware compression pipeline of \method{}. For each target Transformer module category, weight chunks are encoded by a base BSQ codec and a residual BSQ codec, followed by category-wise recovery. After all categories are compressed, a final end-to-end recovery stage can be applied to reduce accumulated cross-category mismatch. All code streams, decoders, optional protected channels, LoRA adapters, and metadata are included in the real storage budget.}
% \label{fig:bisco_pipeline}
% \end{figure*}

\subsection{Unit-Sphere Binary Weight Mapping}
The first design choice of \method{} is to represent a local weight chunk by its direction on a unit hypersphere rather than by an explicit centroid index. For each chunk $\mathbf{x}\in\R^{d}$, the encoder produces a latent vector $\mathbf{z}=E_{\phi}(\mathbf{x})\in\R^{b}$. Following BSQ \cite{zhao2024bsq}, the latent vector is normalized and binarized as
% 中文翻译：BiSCo-LLM 的第一个设计选择是用单位超球面上的方向表示局部权重块，而不是用显式中心向量索引表示。对于每个 d 维权重块，编码器产生 b 维潜变量，随后进行球面归一化和二值化。
\begin{equation}
    \mathbf{u}=\frac{\mathbf{z}}{\|\mathbf{z}\|_2}, \qquad
    \mathbf{q}=\frac{\sign(\mathbf{u})}{\sqrt{b}},
    \quad \mathbf{q}\in\left\{-\frac{1}{\sqrt{b}}, +\frac{1}{\sqrt{b}}\right\}^{b}.
\end{equation}
This mapping imposes two constraints that are relevant to storage-aware LLM weight compression. First, all binary codes have the same $\ell_2$ norm and therefore provide the decoder with a magnitude-normalized discrete input. This reduces the dependence of the decoder on unconstrained latent amplitudes, but does not by itself guarantee lower reconstruction error. Second, the stored code for each chunk is the sign pattern of $\mathbf{q}$ rather than an index into an explicit vector dictionary. Thus, unlike conventional vector quantization with a codebook of $K$ centroids in $\R^d$, the main discrete payload does not require storing $Kd$ floating-point codebook entries. Related codebook-free discrete representation methods, including LFQ and BSQ, have shown that binary latent codes can be trained without an explicit lookup table \cite{yu2024magvitv2,zhao2024bsq}. During training, the non-differentiable sign operation is optimized with a straight-through estimator:
% 中文翻译：该映射引入了两个与存储感知 LLM 权重压缩相关的约束。第一，所有二值码具有相同的 $\ell_2$ 范数，因此为解码器提供幅值归一化的离散输入。这可以降低解码器对不受约束潜变量幅值的依赖，但其本身并不保证更低的重建误差。第二，每个权重块实际存储的是 $\mathbf{q}$ 的符号模式，而不是显式向量字典中的索引。因此，不同于使用 $\R^d$ 中 $K$ 个中心向量的传统向量量化，主离散载荷不需要存储 $Kd$ 个浮点码本条目。包括 LFQ 和 BSQ 在内的相关无码本离散表示方法已经表明，二值潜变量可以在没有显式 lookup table 的情况下进行训练。训练过程中，不可导的 sign 操作使用直通估计器进行优化。
\begin{equation}
    \mathbf{q}_{\mathrm{ste}}=\mathbf{u}+\left(\mathbf{q}-\mathbf{u}\right)_{\mathrm{detach}},
    \qquad
    \widehat{\mathbf{x}}=D_{\theta}(\mathbf{q}_{\mathrm{ste}}).
\end{equation}
% \textcolor{blue}{[Validation experiment: the codebook-free and spherical-code motivation is already supported by LFQ/BSQ literature. The paper only needs to validate its transfer to LLM weights. Compare unconstrained binary latent coding and unit-sphere BSQ on LLaMA-2-7B and Qwen3-8B category-wise replacement. Report reconstruction MSE, bit entropy, WikiText-2 PPL, and real bpw. A compact table is enough; a separate figure is optional.] }

The codec is trained by combining chunk reconstruction, latent commitment, and entropy regularization:
% 中文翻译：编解码器通过组合权重块重建、潜变量 commitment 和熵正则进行训练。
\begin{equation}
    \mathcal{L}_{\mathrm{codec}}
    = \mathcal{L}_{\mathrm{rec}}
    + \lambda_{c}\mathcal{L}_{\mathrm{com}}
    + \lambda_{e}\mathcal{L}_{\mathrm{ent}} .
    \label{eq:codec_loss}
\end{equation}
For a mini-batch of $N$ weight chunks, the three terms are defined as
% 中文翻译：对于一个包含 N 个权重块的 mini-batch，三个损失项定义如下。
\begin{equation}
\begin{aligned}
    \mathcal{L}_{\mathrm{rec}}
    &= \frac{1}{N}\sum_{i=1}^{N}\|\widehat{\mathbf{x}}_i-\mathbf{x}_i\|_2^2, \\
    \mathcal{L}_{\mathrm{com}}
    &= \frac{1}{N}\sum_{i=1}^{N}\|\mathbf{u}_i-\sg(\mathbf{q}_i)\|_2^2, \\
    \mathcal{L}_{\mathrm{ent}}
    &= \frac{1}{N}\sum_{i=1}^{N}\sum_{j=1}^{b}H(\mathbf{p}_{ij})
       - \gamma\sum_{j=1}^{b}H(\bar{\mathbf{p}}_j).
\end{aligned}
\label{eq:codec_loss_terms}
\end{equation}
Here $\sg(\cdot)$ denotes stop-gradient. Following the factorized soft BSQ distribution \cite{zhao2024bsq}, the soft probability of the $j$-th bit is defined over the two spherical code values as
% 中文翻译：其中 sg 表示 stop-gradient。按照 BSQ 的因子化 soft 分布，第 j 个 bit 的 soft 概率定义在两个球面码取值上。
\begin{equation}
    \mathbf{p}_{ij}
    = \left[
    \sigma\left(-\frac{2\tau u_{ij}}{\sqrt{b}}\right),
    \sigma\left(\frac{2\tau u_{ij}}{\sqrt{b}}\right)
    \right], \qquad
    \bar{\mathbf{p}}_j=\frac{1}{N}\sum_{i=1}^{N}\mathbf{p}_{ij}.
\end{equation}
Thus, $\mathcal{L}_{\mathrm{ent}}$ sharpens each soft binary assignment while encouraging batch-level bit diversity. This term is important because an implicit binary code space can be very large, and extreme compression should spend bits on codes that are actually used by LLM weight chunks.
% 中文翻译：因此，L_ent 使每个 soft 二值分配更尖锐，同时鼓励 batch 级别的 bit 多样性。该项很重要，因为隐式二值码空间可能很大，而极低比特压缩应把比特花在真实会被 LLM 权重块使用的码上。

\subsection{Second-Stage Residual BSQ Compression}
The second design choice is to allocate additional bits to the error left by the base spherical codec, rather than to simply enlarge a single binary code. The general idea of additive or residual vector coding is already supported by learned LLM compression methods such as AQLM and VPTQ, where multiple code components or residual quantization improve low-bit approximation \cite{egiazarian2024aqlm,liu2024vptq}. In explicit VQ, increasing the vector dimension or the number of codewords may introduce additional centroid storage and lookup cost. In BSQ, the codebook is implicit, so stored centroids are avoided; nevertheless, a larger implicit binary space is not necessarily an effective use of the available payload when the codec is trained on a finite set of pretrained weight chunks.
% 中文翻译：第二个设计选择是把额外比特分配给基础球面 codec 留下的误差，而不是简单地扩大单个二值码。加性或残差向量编码的一般思想已经得到 AQLM 和 VPTQ 等 LLM 压缩方法的支持，这些方法通过多个码分量或残差量化改善低比特近似。在显式 VQ 中，增大向量维度或码字数量可能引入额外的中心向量存储和查表成本。在 BSQ 中，码本是隐式的，因此可以避免存储中心向量；但是，当 codec 只在有限的预训练权重块集合上训练时，更大的隐式二值空间并不一定能有效利用可用载荷。

\begin{figure}[t]
\centering
\includegraphics[width=0.95\linewidth]{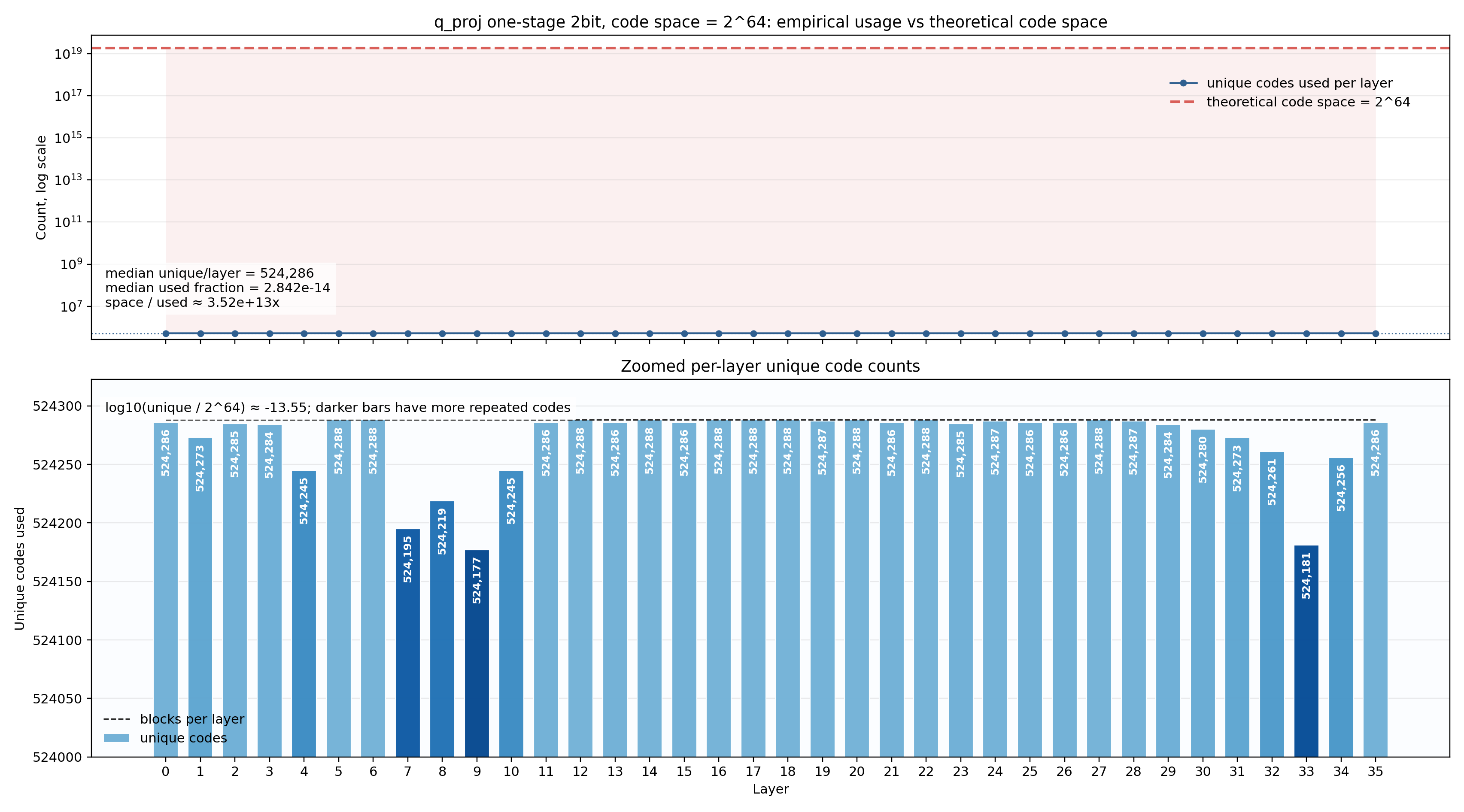}
\caption{Empirical utilization of the implicit BSQ code space on Qwen3-8B $q_{\mathrm{proj}}$ with a one-stage 2-bit setting. The upper panel compares the theoretical binary space $2^{64}$ with the number of occupied code patterns in each layer on a logarithmic scale, where an occupied pattern means that at least one weight chunk is mapped to it. The lower panel zooms into the per-layer occupied-code counts; darker bars indicate layers with more repeated assignments, and the dashed line marks the number of weight chunks per layer. Although most chunks are assigned to distinct code patterns within each layer, the observed occupied-code count remains bounded by the finite number of chunks and covers only a tiny fraction of the theoretical sign-pattern space. This supports the view that simply enlarging a one-stage implicit binary space does not guarantee effective use of the additional representational capacity.}
\label{fig:qproj_code_space_usage}
\end{figure}

To examine this issue, we first measure how much of a large implicit BSQ code space is actually occupied by a fixed pretrained model. We train a one-stage 2-bit BSQ codec for the $q_{\mathrm{proj}}$ matrices of Qwen3-8B and count the number of distinct binary code patterns to which at least one weight chunk is assigned in each layer. In this setting, the binary length is $b=64$, so the theoretical sign-pattern space contains $2^{64}$ possible codes. However, each layer provides only $2^{19}$ weight chunks, which upper-bounds the number of code patterns that can be occupied, observed, and optimized for that layer. As shown in Fig.~\ref{fig:qproj_code_space_usage}, the median number of occupied code patterns per layer is $524{,}286$, corresponding to a used fraction of approximately $2.84\times 10^{-14}$ of the theoretical space. This diagnostic does not imply that BSQ codes are ineffective; instead, it indicates that most sign patterns introduced by a very large one-stage binary space are not occupied by the finite set of pretrained weight chunks.
% 中文翻译：为了考察这个问题，我们首先测量一个大的隐式 BSQ 码空间在固定预训练模型上到底有多少被实际占用。我们针对 Qwen3-8B 的 $q_{\mathrm{proj}}$ 矩阵训练一阶段 2-bit BSQ codec，并统计每一层中至少有一个权重块被映射上去的不同二值 code pattern 数量。在该设置中，二值码长度为 $b=64$，因此理论符号模式空间包含 $2^{64}$ 种可能的 code。然而，每一层只有 $2^{19}$ 个权重块，这从根本上限制了该层可以被占用、观测和优化的 code pattern 数量。如图~\ref{fig:qproj_code_space_usage} 所示，每层被占用 code pattern 数量的中位数为 $524{,}286$，约占理论空间的 $2.84\times 10^{-14}$。这个诊断结果并不表示 BSQ code 无效；它说明的是，由很大一阶段二值空间引入的大多数符号模式，并不会被有限的预训练权重块集合实际占用。

% 中文翻译：图中展示了 Qwen3-8B 的 $q_{\mathrm{proj}}$ 在一阶段 2-bit 设置下对隐式 BSQ 码空间的实际使用情况。上半部分在对数尺度下比较理论二值空间 $2^{64}$ 与每层被占用 code pattern 数量，其中被占用表示至少有一个权重块被映射到该 pattern。下半部分放大展示每层被占用 code 数量；颜色更深的柱表示该层重复映射更多，虚线表示每层权重块数量。虽然每层内部大多数权重块都被分配到不同 code pattern，但可观测到的被占用 code 数量仍然受到有限权重块数量的限制，并且只覆盖理论符号模式空间的极小一部分。这支持了一个观点：单纯扩大一阶段隐式二值空间并不保证能够有效使用额外表示容量。

\begin{figure}[t]
\centering
\includegraphics[width=\columnwidth]{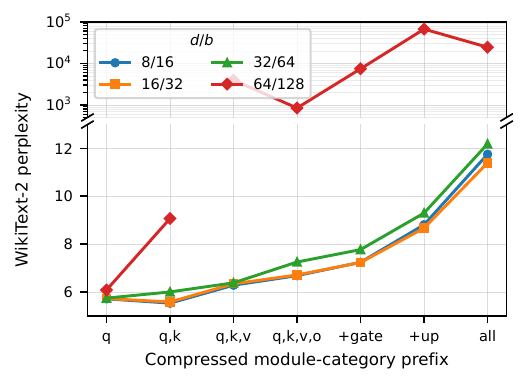}
\caption{Nominal-rate-matched BSQ granularity sweep under category-prefix replacement on LLaMA-2-7B. All settings use the same code-only rate $b/d=2$, while the chunk dimension, binary length, and decoder width increase from $8/16$ to $64/128$. The vertical axis is broken to preserve the trend among stable settings while still showing the instability of the largest granularity. Together with Fig.~\ref{fig:qproj_code_space_usage}, this result motivates assigning extra capacity to a residual BSQ stage rather than only enlarging a single-stage code.}
\label{fig:bsq_granularity_sweep}
\end{figure}

We complement this utilization diagnostic with a nominal-rate-matched granularity sweep on LLaMA-2-7B by replacing an increasing prefix of module categories. The tested settings keep the code-only rate fixed at $b/d=2$, while increasing the chunk dimension $d$, binary length $b$, and decoder width. Thus, this experiment should be interpreted as a practical BSQ granularity diagnostic rather than as an isolated comparison of code-space cardinality. As shown in Fig.~\ref{fig:bsq_granularity_sweep}, small and medium granularities remain stable in the category-prefix replacement test, whereas the largest setting becomes unstable after multiple categories are compressed. The result suggests that, under the current training recipe, naively enlarging a one-stage BSQ code does not provide monotonic gains.
% 中文翻译：我们进一步用 LLaMA-2-7B 上的同名义码率粒度扫描来补充上述使用率诊断，并逐步替换越来越多的模块类别。被测试设置都保持仅码流码率为 $b/d=2$，但同时增大权重块维度 $d$、二值码长度 $b$ 和解码器宽度。因此，这个实验应被理解为实际的 BSQ 粒度诊断，而不是孤立的码空间基数比较。如图~\ref{fig:bsq_granularity_sweep} 所示，小粒度和中等粒度设置在逐类别前缀替换测试中保持稳定，而最大的设置在压缩多个类别后变得不稳定。该结果说明，在当前训练配置下，朴素地扩大一阶段 BSQ 码并不会带来单调增益。

% Author note: the provided source sheet contains paired values in several cells. This figure uses the first value of each pair as the category-prefix WikiText-2 PPL. The meaning of the second value should be confirmed before final submission.
% 中文翻译：图中展示了 LLaMA-2-7B 上逐类别前缀替换设置下的同名义码率 BSQ 粒度扫描。所有设置都使用相同的仅码流码率 $b/d=2$，但权重块维度、二值码长度和解码器宽度从 $8/16$ 增大到 $64/128$。纵轴采用截断形式，以便在保留稳定设置之间趋势的同时展示最大粒度设置的不稳定性。结合图~\ref{fig:qproj_code_space_usage}，该结果说明，与其只扩大单阶段码，不如将额外容量分配给残差 BSQ 阶段。

Taken together, the two diagnostics provide complementary evidence for the residual design. The Qwen3-8B code-space analysis shows that a large implicit binary space can be far larger than the set of code patterns occupied by a fixed group of pretrained weights. The LLaMA-2-7B category-prefix sweep shows that increasing the one-stage granularity under the same nominal code rate does not reliably improve model-level behavior. These observations motivate the base--residual decomposition used by \method{}: the base codec first reconstructs the dominant component, and the residual codec then spends the additional payload on the reconstruction error left by the base stage. This decomposition changes the learning target of the extra bits from the original heterogeneous weight distribution to a lower-energy residual distribution, while keeping each codec stage at a moderate granularity.
% 中文翻译：综合来看，这两个诊断实验为残差设计提供了互补证据。Qwen3-8B 的码空间分析表明，一个大的隐式二值空间可能远大于固定预训练权重集合实际占用的 code pattern 集合。LLaMA-2-7B 的逐类别前缀扫描表明，在相同名义码流码率下增大一阶段粒度，并不能稳定改善模型级行为。这些观察结果为 BiSCo-LLM 采用基础-残差分解提供了动机：基础 codec 先重建主导成分，残差 codec 再把额外载荷用于基础阶段留下的重建误差。这种分解把额外比特的学习目标从原始异质权重分布转移到能量更低的残差分布，同时让每个 codec 阶段保持在适中的粒度。

Let $\mathbf{r}^{(0)}=\mathbf{x}$. The base codec reconstructs $\widehat{\mathbf{r}}^{(1)}$ from the original chunk, and the second-stage codec reconstructs the residual left by the base stage:
% 中文翻译：令初始残差等于原始权重块。基础 codec 从原始块重建第一阶段结果，第二阶段 codec 则重建基础阶段留下的残差。
\begin{align}
    \mathbf{q}^{(1)} &= Q\left(E_{\phi_1}(\mathbf{r}^{(0)})\right), &
    \widehat{\mathbf{r}}^{(1)} &= D_{\theta_1}(\mathbf{q}^{(1)}), \\
    \mathbf{r}^{(1)} &= \mathbf{x}-\widehat{\mathbf{r}}^{(1)}, &
    \mathbf{q}^{(2)} &= Q\left(E_{\phi_2}(\mathbf{r}^{(1)})\right), \\
    \widehat{\mathbf{r}}^{(2)} &= D_{\theta_2}(\mathbf{q}^{(2)}), &
    \widehat{\mathbf{x}} &= \widehat{\mathbf{r}}^{(1)}+\widehat{\mathbf{r}}^{(2)} .
\end{align}
We call this design second-stage residual compression because the main additional capacity is assigned to the first-order reconstruction residual. The formulation can be extended to more stages, but the paper should present the two-stage form as the default setting unless later experiments show that a third stage gives a clearly better real-bpw trade-off.
% 中文翻译：我们称其为二阶残差压缩，因为主要额外容量被分配给一阶重建残差。该形式可以扩展到更多阶段，但除非后续实验表明第三阶段在真实 bpw 下明显更优，正文应以二阶段作为默认设置。

Table~\ref{tab:residual_stage_ppl} provides evidence that adding a residual BSQ stage improves category-wise replacement perplexity in most tested settings. At 32D, the second stage improves all seven categories. The gains are especially visible for MLP projections, where $up\_proj$ improves from 6.23 to 5.95 and $down\_proj$ improves from 6.79 to 6.58. The table also contains a budget-comparable reading: comparing one-stage 32D BSQ with two-stage 16D$\times$2 BSQ gives the same total binary code length, and the two-stage form improves six out of seven categories. These results support the claim that the residual left by the base codec still contains recoverable information useful for model behavior.
% 中文翻译：表格提供了证据：加入残差 BSQ 阶段在多数设置下改善逐类别替换困惑度。在 32D 下，二阶阶段改善全部七个类别，其中 MLP projection 收益明显。该表还包含一个近似预算对齐的读法：一阶段 32D BSQ 与二阶段 16D$\times$2 BSQ 具有相同二值码长度，二阶段形式在七个类别中改善六个。这说明基础 codec 留下的残差中仍包含对模型行为有用的可恢复信息。

\begin{table}[t]
\centering
\caption{Perplexity of LLaMA-2-7B when compressing one linear category at a time with one-stage and two-stage BSQ. Lower is better. $\Delta$ is two-stage PPL minus one-stage PPL.}
\label{tab:residual_stage_ppl}
\footnotesize
\setlength{\tabcolsep}{2.2pt}
\begin{tabular}{lcccccc}
\toprule
Category & 16D & 16D$\times$2 & $\Delta_{16}$ & 32D & 32D$\times$2 & $\Delta_{32}$ \\
\midrule
$q\_proj$    & 5.72 & \textbf{5.65} & -0.07 & 5.74 & \textbf{5.67} & -0.07 \\
$k\_proj$    & 5.69 & \textbf{5.66} & -0.03 & 5.86 & \textbf{5.69} & -0.17 \\
$v\_proj$    & \textbf{6.12} & 6.18 & +0.06 & 6.11 & \textbf{6.05} & -0.06 \\
$o\_proj$    & 5.88 & \textbf{5.84} & -0.04 & 5.92 & \textbf{5.78} & -0.14 \\
$gate\_proj$ & 6.09 & \textbf{6.00} & -0.09 & 6.15 & \textbf{5.95} & -0.20 \\
$up\_proj$   & 6.24 & \textbf{6.03} & -0.21 & 6.23 & \textbf{5.95} & -0.28 \\
$down\_proj$ & 6.65 & \textbf{6.51} & -0.14 & 6.79 & \textbf{6.58} & -0.21 \\
\bottomrule
\end{tabular}
\end{table}
% 中文翻译：表中展示了在每次只压缩一个 linear 类别时，一阶段和二阶段 BSQ 在 LLaMA-2-7B 上的困惑度。数值越低越好。Delta 表示二阶段 PPL 减去一阶段 PPL，因此负值表示二阶段残差编码带来改进。

\subsection{Auxiliary 8-Bit Sensitive-Channel Path}
The base and residual BSQ codecs define the primary compressed representation of \method{}. However, at an extreme 2-bit budget, applying the same BSQ representation to all channels is unnecessarily risky. Prior studies on LLM quantization have shown that a small number of activation-salient or outlier-associated channels can dominate low-bit degradation \cite{dettmers2022llmint8,lin2023awq,dettmers2023spqr,kim2023squeezellm}. We therefore first identify a small set of high-impact channels before BSQ compression, store the corresponding original slices with an 8-bit quantizer, and apply the two-stage BSQ codecs only to the remaining unprotected slices. This auxiliary path is not intended to replace the BSQ representation, and its storage is counted separately in $S_{\mathrm{8bit}}$.
% 中文翻译：基础 BSQ codec 和残差 BSQ codec 构成 BiSCo-LLM 的主要压缩表示。然而，在极限 2-bit 预算下，对所有通道使用相同的 BSQ 表示会带来不必要的风险。已有 LLM 量化研究表明，少量激活显著或与离群值相关的通道可能主导低比特退化。因此，我们先在 BSQ 压缩之前识别少量高影响通道，将对应原始切片用 8bit 量化器存储，并且只对未保护的剩余切片应用二阶段 BSQ codec。该辅助路径并不替代 BSQ 表示，其存储开销单独计入 S_8bit。

The protected set is selected from the original weight and calibration statistics rather than from a BSQ reconstruction residual. For attention projections, we protect input channels. With $\mathbf{W}_{\ell}^{c}\in\mathbb{R}^{d_{\mathrm{out}}\times d_{\mathrm{in}}}$, an input channel corresponds to a column slice. We score column $j$ by combining an activation-dependent statistic $g_{\ell,j}^{c}$, such as mean absolute activation, with the magnitude of the corresponding original weight slice:
% 中文翻译：受保护集合是根据原始权重和校准统计选择的，而不是根据 BSQ 重建残差选择的。对于注意力投影，我们保护输入通道。若 W 的形状为输出维度乘输入维度，则输入通道对应列切片。我们用激活相关统计量 g（例如平均绝对激活值）与对应原始权重切片幅值相乘，得到列 j 的分数：
\begin{equation}
    S_{\ell,j}^{c}=g_{\ell,j}^{c}\left\|\mathbf{W}_{\ell,:,j}^{c}\right\|_2,
    \qquad
    \mathcal{P}_{\ell}^{c}=\TopK_j\left(S_{\ell,j}^{c},\left\lceil \rho d_{\mathrm{in}}\right\rceil\right),
    \label{eq:protected_channel_selection}
\end{equation}
where $c\in\{q\_proj,k\_proj,v\_proj,o\_proj\}$ and $\rho$ is the protected-channel ratio.
% 中文翻译：其中 c 属于 q_proj、k_proj、v_proj 和 o_proj，rho 表示受保护通道比例。

For gated MLP blocks, the intermediate channel is shared by $gate\_proj$, $up\_proj$, and $down\_proj$. Protecting these matrices independently can break this structural coupling. We therefore use coupled intermediate-channel protection before BSQ compression: the same intermediate index is selected for the output rows of $gate\_proj$ and $up\_proj$ and the corresponding input column of $down\_proj$. Following this coupled-channel view, the MLP score is aggregated as
% 中文翻译：对于门控 MLP block，中间通道由 gate_proj、up_proj 和 down_proj 共享。如果三个矩阵独立选择保护通道，可能破坏这种结构耦合。因此，我们在 BSQ 压缩之前使用耦合中间通道保护：同一个中间维度索引会同时用于 gate_proj 和 up_proj 的输出行，以及 down_proj 中对应的输入列。按照这种耦合通道视角，MLP 分数聚合为：
\begin{equation}
\begin{aligned}
    S_{\ell,j}^{\mathrm{mlp}}=
    &\;g_{\ell,j}^{gate}\left\|\mathbf{W}_{\ell,j,:}^{gate}\right\|_2
    +g_{\ell,j}^{up}\left\|\mathbf{W}_{\ell,j,:}^{up}\right\|_2 \\
    &+g_{\ell,j}^{down}\left\|\mathbf{W}_{\ell,:,j}^{down}\right\|_2, \\
    \mathcal{P}_{\ell}^{\mathrm{mlp}}
    &=\TopK_j\left(S_{\ell,j}^{\mathrm{mlp}},\left\lceil \rho d_{\mathrm{ffn}}\right\rceil\right).
\end{aligned}
\label{eq:mlp_coupled_protection}
\end{equation}
This coupled selection follows the same rationale as SpenseGPT-style aligned MLP-channel selection, where $gate\_proj$, $up\_proj$, and $down\_proj$ are treated as a coupled structure rather than as independent matrices~\cite{lee2026spensegpt}.
% 中文翻译：这种耦合选择与 SpenseGPT 式 MLP 对齐通道选择的思想一致，即 gate_proj、up_proj 和 down_proj 被视为一个耦合结构，而不是彼此独立的矩阵。

After the protected set is fixed, the selected slices are stored with an 8-bit quantizer. The BSQ codec is trained and applied on the complement, and the final reconstructed matrix is assembled as
% 中文翻译：受保护集合确定后，被选中切片使用 8bit 量化器存储。BSQ codec 在补集上训练和应用，最终重建矩阵组装为：
\begin{equation}
    \widehat{\mathbf{W}}_{\ell}^{c}[\mathcal{S}_{\ell}^{c}]
    =Q_8\left(\mathbf{W}_{\ell}^{c}[\mathcal{S}_{\ell}^{c}]\right),
    \qquad
    \widehat{\mathbf{W}}_{\ell}^{c}[\overline{\mathcal{S}}_{\ell}^{c}]
    =\widehat{\mathbf{W}}_{\ell,\mathrm{BSQ}}^{c}[\overline{\mathcal{S}}_{\ell}^{c}],
    \label{eq:protected_8bit_scatter}
\end{equation}
where $\mathcal{S}_{\ell}^{c}$ denotes the selected column, row, or coupled MLP slices induced by $\mathcal{P}_{\ell}^{c}$, and $Q_8(\cdot)$ is a standard per-channel or per-group 8-bit quantizer.
% 中文翻译：其中 S 表示由 P 导出的被选中列切片、行切片或耦合 MLP 切片，Q8 表示标准 per-channel 或 per-group 8bit 量化器。

The usefulness of this auxiliary path is supported by the category-replacement diagnostic in Table~\ref{tab:prot_effectiveness}. Each experiment replaces one module category and evaluates the full model after replacement. In the protected variant, we first select and store 1\% protected slices in 8-bit, then train and apply the same two-stage 2-bit BSQ setting on the remaining slices. Attention projections use input-channel protection, and MLP projections use coupled intermediate-channel protection. With this small protected set, the average WikiText-2 perplexity decreases from 5.95 to 5.82, and all seven categories improve over the unprotected baseline. The largest gain appears on $down_{\mathrm{proj}}$, where the perplexity decreases from 6.58 to 6.10. These results motivate the use of protected channels as a lightweight safety mechanism in the extreme low-bit setting. The protection axis and protection ratio are further analyzed in the ablation section.
% 中文翻译：表中的逐类别替换诊断实验支持了该辅助路径的有效性。每个实验替换一个模块类别，并在替换后评估完整模型。在保护版本中，我们先选择 1\% 受保护切片并用 8bit 存储，然后在剩余切片上训练和应用相同的二阶段 2-bit BSQ 设置。注意力投影使用输入通道保护，MLP 投影使用耦合中间通道保护。使用这一小比例保护集合后，WikiText-2 平均困惑度从 5.95 降低到 5.82，并且七个类别都优于无保护基线。最大收益出现在 down_proj 上，困惑度从 6.58 降低到 6.10。这些结果说明，在极低比特设置下，重要通道保护可以作为轻量安全机制。保护轴和保护比例会在消融部分进一步分析。

\begin{table}[t]
\centering
\caption{Effectiveness of 1\% protected channels under 2-bit BSQ category replacement on LLaMA-2-7B. WikiText-2 perplexity is reported, and lower is better.}
\label{tab:prot_effectiveness}
\footnotesize
\setlength{\tabcolsep}{4.5pt}
\begin{tabular}{lccc}
\toprule
Category & No Prot. & Prot. 1\% & $\Delta$ \\
\midrule
$q_{\mathrm{proj}}$    & 5.67 & \textbf{5.59} & -0.08 \\
$k_{\mathrm{proj}}$    & 5.69 & \textbf{5.58} & -0.11 \\
$v_{\mathrm{proj}}$    & 6.05 & \textbf{5.86} & -0.19 \\
$o_{\mathrm{proj}}$    & 5.78 & \textbf{5.76} & -0.02 \\
$gate_{\mathrm{proj}}$ & 5.95 & \textbf{5.94} & -0.01 \\
$up_{\mathrm{proj}}$   & 5.95 & \textbf{5.92} & -0.03 \\
$down_{\mathrm{proj}}$ & 6.58 & \textbf{6.10} & -0.48 \\
\midrule
Avg. & 5.95 & \textbf{5.82} & -0.13 \\
\bottomrule
\end{tabular}
\end{table}
% 中文翻译：表中展示了 LLaMA-2-7B 上二阶段 2-bit BSQ 逐类别替换时，1\% 保护通道的有效性。数值为 WikiText-2 困惑度，越低越好。

\subsection{Category-Wise Recovery Distillation}
Although the BSQ codecs are trained to minimize weight reconstruction error, reconstruction alone is not sufficient for preserving the final model behavior under extreme low-bit compression. Even small weight perturbations can be amplified through activations, residual connections, normalization layers, and subsequent Transformer blocks, and the errors introduced by different compressed categories can accumulate into an unacceptable task-level degradation. Layer-wise or local distillation has already been used in LLM quantization systems such as ZeroQuant, and low-rank adapters are a standard parameter-efficient recovery mechanism after quantization \cite{yao2022zeroquant,hu2021lora,dettmers2023qlora}. However, a layer-wise recovery objective remains local: it can reduce the loss measured at the current block while still failing to recover the behavior of the assembled compressed model. Therefore, after compressing and replacing one module category, \method{} performs category-wise recovery distillation before moving to the next category. The teacher is the original model or the current reference model before this category replacement, and the student is the partially compressed model after replacement.
% 中文翻译：虽然 BSQ codec 通过最小化权重重建误差进行训练，但在极低比特压缩下，仅依靠重建优化不足以保持最终模型行为。即使很小的权重扰动也可能经过激活、残差连接、归一化层和后续 Transformer block 被放大，而不同压缩类别引入的误差还会累积成不可接受的任务级性能下降。逐层或局部蒸馏已被用于 ZeroQuant 等 LLM 量化系统，低秩 adapter 也是量化后参数高效恢复的常用机制。然而，逐层恢复目标本质上仍然是局部目标：它可以降低当前 block 上测得的损失，但仍然无法保证组装后的压缩模型行为被恢复。因此，每压缩并替换一个模块类别后，BiSCo-LLM 在进入下一个类别前执行逐类别恢复蒸馏。教师模型为原始模型或当前类别替换前的参考模型，学生模型为替换该类别后的部分压缩模型。

\begin{figure*}[t]
\centering
\begin{minipage}[t]{0.49\textwidth}
\centering
\includegraphics[width=\linewidth]{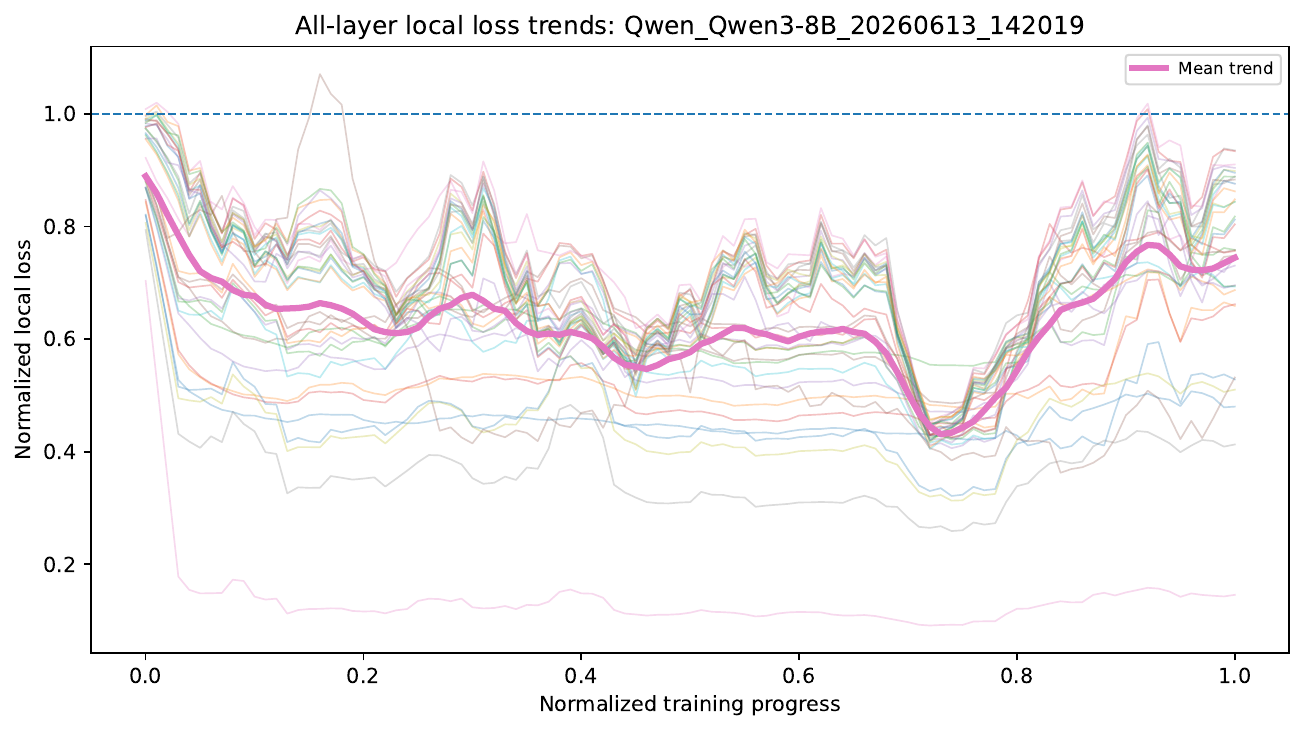}
\end{minipage}
\hfill
\begin{minipage}[t]{0.49\textwidth}
\centering
\includegraphics[width=\linewidth]{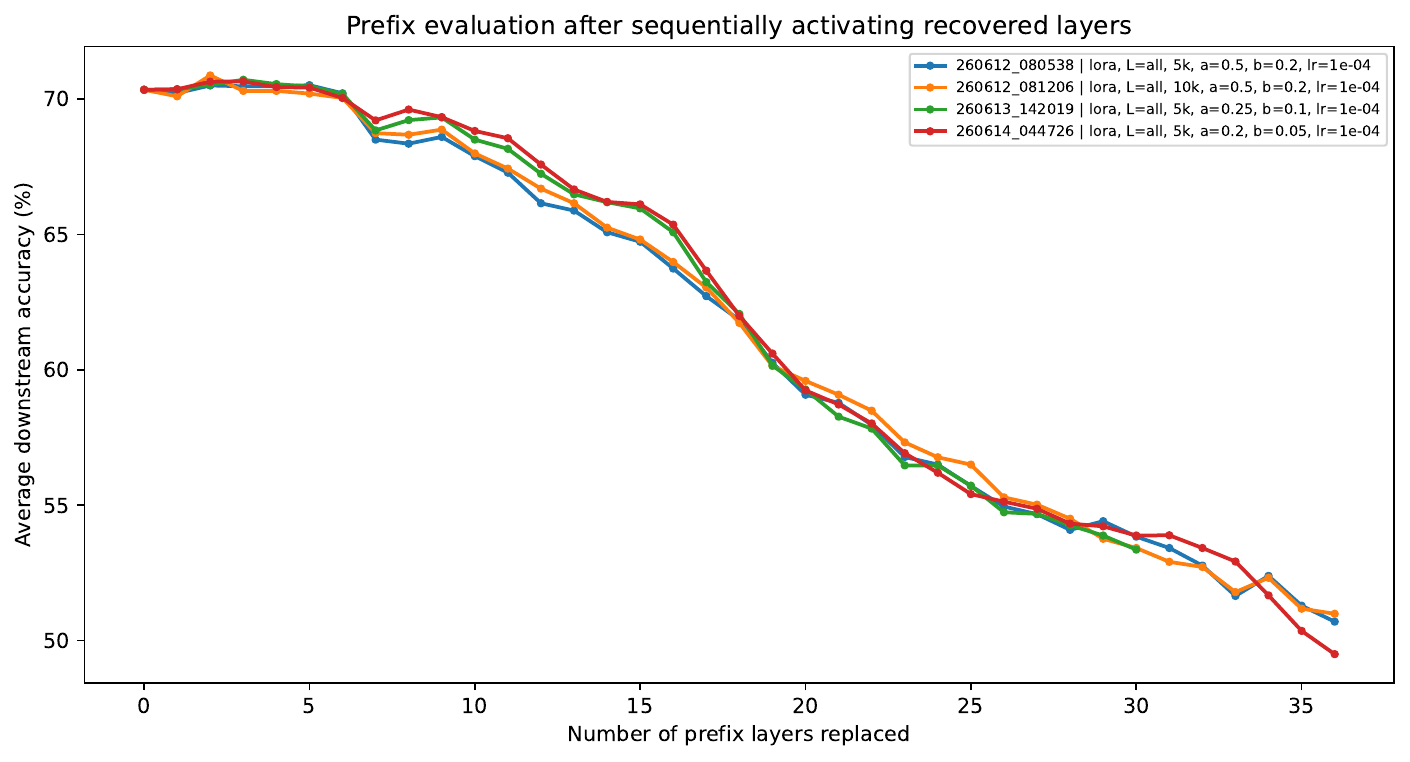}
\end{minipage}
\caption{Diagnostic comparison between layer-wise local convergence and prefix-level model behavior on Qwen3-8B. Left: all-layer normalized local losses for the layer-wise recovery setting $\alpha=0.25$, $\beta=0.1$, 5k steps, and learning rate $10^{-4}$. Thin curves denote individual blocks and the thick curve denotes the mean trend. Right: prefix evaluation when recovered layers are progressively activated; the corresponding curve is included among the evaluated hyper-parameter settings. The matched setting reduces local losses for all parsed blocks, but the downstream accuracy still degrades at later prefix points. This result is used as a diagnostic motivation for category-wise recovery rather than as an exhaustive negative result for all possible layer-wise recovery configurations.}
\label{fig:layerwise_recovery_mismatch}
\end{figure*}

Fig.~\ref{fig:layerwise_recovery_mismatch} provides a diagnostic example on Qwen3-8B. In this experiment, each Transformer block is recovered with a layer-wise LoRA objective for 5k steps, using the setting $\alpha=0.25$, $\beta=0.1$, and learning rate $10^{-4}$. The local objective is not under-optimized in this run: all 36 block-level curves finish below their initial values, and the mean first-to-last loss reduction is 27.26\%. Nevertheless, the matched prefix evaluation does not show reliable full-model recovery. The average downstream accuracy is 70.34\% before activating recovered layers, reaches 70.71\% at the best early prefix point, and decreases to 53.36\% at the last evaluated prefix point. This observation does not prove that all layer-wise recovery strategies must fail, but it indicates that local convergence alone is not sufficient evidence of global recovery. It motivates the category-wise schedule used in \method{}, where recovery is performed after a coherent module category has been replaced and evaluated in the context of the remaining model.
% 中文翻译：图~\ref{fig:layerwise_recovery_mismatch} 给出了 Qwen3-8B 上的一个诊断例子。在该实验中，每个 Transformer block 使用逐层 LoRA 目标恢复 5k 步，设置为 $\alpha=0.25$、$\beta=0.1$、学习率 $10^{-4}$。该 run 中局部目标并不是没有优化：36 个 block 级曲线最终都低于初始值，首末平均损失下降为 27.26\%。然而，与之匹配的 prefix evaluation 并没有显示可靠的整模型恢复。激活恢复层之前，平均下游精度为 70.34\%；早期 prefix 的最佳点为 70.71\%；最后一个被评测的 prefix 点下降到 53.36\%。这个观察并不能证明所有逐层恢复策略都一定失败，但它说明局部收敛本身不足以作为全局恢复的证据。这为 BiSCo-LLM 使用逐类别恢复流程提供了动机：恢复应在一个结构一致的模块类别被替换后，在剩余模型的上下文中进行。

% 中文翻译：该图比较了 Qwen3-8B 上逐层局部收敛与 prefix 级模型行为。左图为逐层恢复设置 $\alpha=0.25$、$\beta=0.1$、5k 步、学习率 $10^{-4}$ 下所有层的归一化局部损失；细线表示单个 block，粗线表示平均趋势。右图为逐步激活恢复层时的 prefix evaluation；其中包含与左图设置对应的曲线。匹配设置下所有已解析 block 的局部损失都下降，但后期 prefix 点的下游精度仍然下降。该结果被用作逐类别恢复的诊断性动机，而不是对所有可能逐层恢复配置的穷尽性否定结论。
% \textcolor{blue}{[Before submission, verify the checkpoint-to-run mapping between Qwen\_Qwen3-8B\_20260613\_142019 and the prefix-evaluation summary, because the raw prefix logs store checkpoint paths rather than this run identifier.]}

The recovery objective can combine output distribution distillation and hidden-state alignment:
% 中文翻译：恢复目标可以结合输出分布蒸馏和 hidden-state 对齐。
\begin{equation}
    \mathcal{L}_{\mathrm{recover}}
    =\lambda_{\mathrm{kd}}\mathcal{L}_{\mathrm{KD}}
    +\lambda_{\mathrm{hid}}\mathcal{L}_{\mathrm{hidden}}
    +\lambda_{\mathrm{aux}}\mathcal{L}_{\mathrm{aux}}.
\end{equation}
In the default setting, only low-rank adapters are trained during recovery, while the BSQ codes are fixed. This makes the recovery payload explicit and prevents the distillation stage from hiding compression cost inside full-precision weight updates. The implementation supports several recovery placements: LoRA on the compressed category, LoRA on the remaining uncompressed categories, decoder tuning, and decoder-plus-LoRA tuning. 
% 中文翻译：默认设置中，恢复阶段只训练低秩 adapter，BSQ 码固定不动。这样恢复载荷是显式的，也避免蒸馏阶段通过全精度权重更新隐藏压缩成本。实现上支持多种恢复位置，包括在被压缩类别上加 LoRA、在未压缩类别上加 LoRA、调节 decoder，以及 decoder 加 LoRA 联合调节。

After all target categories are compressed, an optional end-to-end LoRA fine-tuning stage can be applied. This final stage addresses accumulated cross-category mismatch, while category-wise recovery addresses the local mismatch immediately after each replacement. 
% 中文翻译：所有目标类别压缩完成后，可以使用可选端到端 LoRA 微调。最终阶段处理跨类别误差累积，逐类别恢复则处理每次替换后的局部错配。

\subsection{Engineering Optimizations}
To improve the usability and practical value of the proposed compression framework, we further incorporate a set of engineering optimizations around training scalability and deployment efficiency. Specifically, category-batched codec optimization jointly processes the same module category across layers as parallel training instances, while batched parallel residual decoding avoids a sequential loading bottleneck for multi-stage reconstruction. These optimizations are important for applying \method{} to LLM-scale models, although they do not change the mathematical form of the compressed representation or the storage accounting.
% 中文翻译：为了提升所提出压缩框架的可用性和实际价值，我们进一步围绕训练可扩展性和部署效率加入了一系列工程优化。具体而言，类别 batch 化 codec 优化将同一模块类别在不同层中的权重作为并行训练实例共同处理，而 batch 并行残差解码避免了多阶段重建在加载阶段形成顺序瓶颈。这些优化对于将 BiSCo-LLM 应用于 LLM 规模模型非常重要，但不改变压缩表示的数学形式，也不改变存储开销的核算方式。

\paragraph{Category-batched codec optimization.}
Optimizing one BSQ codec for each individual linear layer would be unnecessarily expensive. In Transformer LLMs, linears from the same module category, such as all $down\_proj$ layers, usually have identical tensor shapes across layers. We exploit this structure by collecting the weight chunks of all layers in the same category and treating them as parallel training instances of a shared codec optimization. Concretely, the prepared per-layer tensors are stacked along a model dimension and reshaped into batches of weight chunks, so a single optimization loop can update the category codec while processing multiple layers in parallel. This training-time batching improves codec optimization efficiency without introducing an additional stored representation; the resulting model still explicitly stores the BSQ code streams, decoders, optional 8-bit payloads, and metadata.
% 中文翻译：如果为每个单独 linear 层逐个优化一个 BSQ codec，训练成本会过高。在 Transformer LLM 中，同一模块类别的线性层，例如所有 down_proj 层，通常在不同层之间具有相同张量形状。我们利用这一结构，收集同一类别中所有层的权重块，并将它们作为共享 codec 优化中的并行训练实例。具体而言，预处理后的逐层张量会沿 model 维度堆叠，并重新组织成权重块 batch，使单个优化循环能够在处理多层权重的同时更新类别 codec。该训练阶段 batch 化提升了 codec 优化效率，但不引入额外存储表示；最终模型仍然显式存储 BSQ 码流、解码器、可选 8bit 载荷和元数据。

\paragraph{Batched parallel residual decoding.}
The residual dependency between BSQ stages is used to define the training target of the second stage, but it does not require sequential decoding after training. Once the binary code tensors and decoders are fixed, deployment can decode the base and residual stages independently and sum the reconstructed tensors:
% 中文翻译：BSQ 阶段之间的残差依赖用于定义第二阶段的训练目标，但训练完成后并不要求顺序解码。一旦二值码张量和解码器固定，部署时可以独立解码基础阶段和残差阶段，并将重建张量相加。
\begin{equation}
    \widehat{\mathbf{W}}
    = \Assemble\left(D_{\theta_1}(\mathbf{B}^{(1)})\right)
    + \Assemble\left(D_{\theta_2}(\mathbf{B}^{(2)})\right).
    \label{eq:parallel_stage_decode}
\end{equation}
When the stage decoders share the same architecture, we merge the stage index into the batch or group dimension and perform a packed decoding pass. The decoded tensors are then reshaped to recover the stage dimension and reduced by summation. This implements multi-stage residual decoding as batched parallel decoding and makes the decoding overhead measurable separately from accuracy.
% 中文翻译：当不同阶段的 decoder 结构相同时，我们将 stage 索引合并到 batch 或 group 维度中，并执行一次打包解码。解码后的张量随后恢复 stage 维度并求和。这将多级残差解码实现为 batch 并行解码，并使解码开销可以与精度指标分开度量。

% \textcolor{blue}{[Experiment design for engineering cost: report codec training wall-clock time with and without category-parallel scheduling, and report load-time decoding latency for sequential versus packed residual decoding. These results should be presented as implementation efficiency results, not as core algorithmic ablations.]}
% 中文翻译：工程开销实验应报告有无类别并行调度时的 codec 训练耗时，以及顺序残差解码与打包残差解码的加载延迟。这些结果应作为实现效率结果展示，而不是核心算法消融。

% \textcolor{blue}{[Figure design: if this part is illustrated, use a small implementation diagram: one panel for category-parallel BSQ training across GPUs and one panel for packed residual decoding. The caption should state that these are engineering optimizations and are not counted as separate algorithmic components.]}

\section{Experiments}\label{sec:experiments}
\subsection{Experimental Setup}
\textbf{Models.} We plan to evaluate \method{} on representative open LLM families, including LLaMA-3 8B, Qwen3 8B. For each model, we will report the original FP16/BF16 performance and several compressed variants under different real bit budgets.
% 中文翻译：模型方面，我们计划在代表性的开源 LLM 家族上评估 BiSCo-LLM，包括 LLaMA-2/3、Qwen2.5/Qwen3、Mistral 和其他模型。对于每个模型，我们会报告原始 FP16/BF16 性能以及不同真实比特预算下的若干压缩版本。

\textbf{Target modules.} Unless otherwise specified, the target linear categories are $q\_proj$, $k\_proj$, $v\_proj$, $o\_proj$, $gate\_proj$, $up\_proj$, and $down\_proj$. We will also evaluate partial compression settings, such as compressing only $down\_proj$ or only attention projections, to analyze category sensitivity.
% 中文翻译：目标模块方面，除非另有说明，目标线性层类别包括 q_proj、k_proj、v_proj、o_proj、gate_proj、up_proj 和 down_proj。我们也会评估局部压缩设置，例如只压缩 down_proj 或只压缩注意力投影，以分析类别敏感性。

\textbf{Calibration and distillation data.} The codec is optimized mainly on weight chunks, whereas calibration samples are used to estimate activation-dependent statistics for activation-aware reconstruction objectives and sensitivity-based channel selection. For recovery training, we follow the quantization-aware distillation setting of EdgeRazor~\cite{zhang2026edgerazor} and construct a mixed distillation corpus from instruction-following, reasoning, educational, reading-comprehension, science-question answering, and long-context data sources. The mixture includes Alpaca~\cite{taori2023alpaca}, OpenOrca~\cite{lian2023openorca}, FineWeb-Edu~\cite{penedo2024fineweb}, RACE~\cite{lai2017race}, SciQ~\cite{welbl2017sciq}, LongAlpaca~\cite{chen2023longlora}, and LongAlign~\cite{bai2024longalign}. Alpaca and OpenOrca provide general instruction-following and teacher-generated reasoning examples, FineWeb-Edu provides high-quality educational text, RACE and SciQ introduce reading-comprehension and science-question answering samples, and LongAlpaca and LongAlign improve the coverage of long-context instruction-following cases. Unless otherwise specified, the same data mixture is used for category-wise recovery distillation and the final end-to-end distillation stage. For reproducibility, we report the number of sampled examples from each source, the sampling ratio, the total number of tokens, the maximum sequence length, the tokenizer, the random seed, and all preprocessing rules. We also keep the calibration and distillation data disjoint from evaluation benchmarks whenever the original dataset splits make such separation possible.
% 中文翻译：校准和蒸馏数据方面，codec 主要在权重块上进行优化，而校准样本用于估计与激活相关的统计信息，以支持激活感知重建目标和基于敏感性的通道选择。对于恢复训练，我们遵循 EdgeRazor 的量化感知蒸馏设置，并从指令跟随、推理、教育文本、阅读理解、科学问答和长上下文数据源中构造混合蒸馏语料。该混合数据包括 Alpaca、OpenOrca、FineWeb-Edu、RACE、SciQ、LongAlpaca 和 LongAlign。Alpaca 和 OpenOrca 提供通用指令跟随与教师生成的推理样本，FineWeb-Edu 提供高质量教育文本，RACE 和 SciQ 引入阅读理解与科学问答样本，LongAlpaca 和 LongAlign 提升长上下文指令跟随样本的覆盖。除非另有说明，逐类别恢复蒸馏和最终端到端蒸馏阶段使用相同的数据混合。为保证可复现性，我们报告每个数据源的采样样本数量、采样比例、总 token 数、最大序列长度、tokenizer、随机种子和所有预处理规则。只要原始数据集划分允许，我们也会保证校准/蒸馏数据与评估基准不重叠。

\textbf{Evaluation tasks.} We report perplexity on WikiText-2~\cite{merity2017wikitext} and C4~\cite{raffel2020t5}, and zero-shot or few-shot accuracy on common language-understanding benchmarks, including BoolQ~\cite{clark2019boolq}, RTE from GLUE~\cite{wang2018glue}, WinoGrande~\cite{sakaguchi2020winogrande}, ARC-Easy and ARC-Challenge~\cite{clark2018arc}, OpenBookQA~\cite{mihaylov2018openbookqa}, PIQA~\cite{bisk2020piqa}, and MMLU~\cite{hendrycks2021mmlu}.
% For reasoning-oriented analysis, we will additionally consider \todo{GSM8K, MATH, BBH, or other reasoning tasks if time allows}.
% 中文翻译：评估任务方面，我们会报告 WikiText-2 和 C4 上的困惑度，并在常用语言理解基准上报告 zero-shot 或 few-shot 精度，包括 BoolQ、GLUE 中的 RTE、WinoGrande、ARC-Easy、ARC-Challenge、OpenBookQA、PIQA 和 MMLU。
% 对于推理分析，如果时间允许，我们还会考虑 GSM8K、MATH、BBH 或其他推理任务。

\subsection{Comparison Baselines}
The comparison baselines are selected according to the methods reported in Table~\ref{tab:main_results} and are organized by quantization granularity. The scalar or group-wise PTQ group includes GPTQ~\cite{frantar2022gptq}, SpinQuant~\cite{liu2024spinquant}, OSTQuant~\cite{hu2025ostquant}, QuIP~\cite{chee2023quip}, and AWQ~\cite{lin2023awq} when low-bit compatible results are available. These methods represent strong scalar-rounding or group-wise quantization pipelines with Hessian reconstruction, activation-aware calibration, incoherence processing, or equivalent rotation/scaling transformations.
% 中文翻译：对比基线按照表~\ref{tab:main_results} 中报告的方法选择，并按量化粒度组织。标量或分组 PTQ 组包括 GPTQ、SpinQuant、OSTQuant、QuIP，以及在低比特设置下有可比结果时的 AWQ。这些方法代表了强标量舍入或分组量化流程，包含 Hessian 重建、激活感知校准、非相干处理，或等价旋转/缩放变换等技术。

The vector or structured-code group includes AQLM~\cite{egiazarian2024aqlm}, VPTQ~\cite{liu2024vptq}, QuIP\#~\cite{tseng2024quipsharp}, LiftQuant~\cite{he2026liftquant}, and UniSVQ~\cite{wang2026unisvq}. These baselines are the most directly related to \method{}, because they reduce low-bit quantization error using trellis codes, additive or vector quantization, lattice codebooks, dimensional lifting and projection, or unified scalar-vector coding. For a fair storage comparison, we count all auxiliary components required by each method, including codebooks, projection matrices, scaling factors, residual payloads, and metadata, when computing the effective bit-width.
% QTIP~\cite{tseng2024qtip}, 
% 中文翻译：向量或结构化编码组包括 QTIP、AQLM、VPTQ、QuIP#、LiftQuant 和 UniSVQ。这些基线与 BiSCo-LLM 最直接相关，因为它们通过 trellis code、加性或向量量化、格码本、维度提升与投影，或统一标量-向量编码来降低低比特量化误差。为了进行公平的存储比较，在计算有效 bit-width 时，我们统计每种方法所需的全部辅助组件，包括码本、投影矩阵、缩放因子、残差载荷和元数据。

\subsection{Main Comparison Results}

\begin{table*}[t]
\centering
\caption{Main comparison of 2-bit quantization methods on Qwen3-8B and LLaMA3-8B. C4 perplexity, RTE, OpenBookQA, MMLU, and bit-width columns are omitted for compactness. Avg. follows the available evaluation set reported by each method; for \method{} on Qwen3-8B, it is computed over BoolQ, RTE, WinoGrande, ARC-Easy, ARC-Challenge, OpenBookQA, and PIQA, while the LLaMA3-8B full-precision Avg. follows the reported set of WinoGrande, PIQA, OpenBookQA, ARC-Easy, and ARC-Challenge.}
\label{tab:main_results}
\scriptsize
\setlength{\tabcolsep}{3.8pt}
\begin{tabular}{lllcccccccc}
\toprule
Model & Type & Method & Wiki$\downarrow$ & AC$\uparrow$ & AE$\uparrow$ & BQ$\uparrow$ & HS$\uparrow$ & PQ$\uparrow$ & WG$\uparrow$ & Avg.$\uparrow$ \\
\midrule
\multirow{9}{*}{Qwen3-8B}
& FP16/BF16               & Full precision & 9.73                  & 56.40 & 80.89 & 86.64 & --    & 77.48 & 68.43 & 69.92 \\
\cmidrule(lr){2-11}
& \multirow{4}{*}{Scalar} & GPTQ           & $4.68{\times}10^{4}$ & 26.79 & 25.67 & 42.87 & 25.84 & 52.50 & 50.04 & 37.29 \\
&                         & SpinQuant      & 17.82                 & 30.46 & 48.40 & 67.06 & 47.94 & 63.17 & 58.64 & 52.61 \\
&                         & OSTQuant       & 26.08                 & 34.64 & 60.02 & 73.21 & 50.29 & 68.50 & 58.17 & 57.47 \\
&                         & QuIP           & 27.61                 & 24.91 & 31.69 & 54.89 & 41.41 & 59.90 & 50.12 & 43.82 \\
\cmidrule(lr){2-11}
& \multirow{3}{*}{Vector} & AQLM           & 18.26                 & 45.22 & 72.31 & 73.73 & 60.99 & 73.07 & 64.33 & 64.94 \\
&                         & QuIP\#         & 12.37                 & 46.50 & 68.43 & 83.12 & 66.62 & 74.32 & 66.30 & 67.55 \\
&                         & UniSVQ         & 14.82                 & 45.82 & 72.35 & 85.07 & 63.18 & 74.16 & 67.09 & 67.94 \\
\cmidrule(lr){2-11}
& Ours                   & \method{}      & 10.18                 & 51.88 & 78.87 & 82.20 & --    & 76.17 & 67.64 & 68.05 \\
\midrule
\multirow{12}{*}{LLaMA3-8B}
& FP16/BF16               & Full precision & 6.14                  & 53.15 & 77.56 & --    & --    & 80.57 & 73.16 & 65.85 \\
\cmidrule(lr){2-11}
& \multirow{5}{*}{Scalar} & GPTQ           & $2.55{\times}10^{6}$ & 25.93 & 25.08 & 46.20 & 26.27 & 51.63 & 49.88 & 37.50 \\
&                         & SpinQuant      & 27.60                 & 21.41 & 33.45 & 60.73 & 39.93 & 56.20 & 55.16 & 44.48 \\
&                         & OSTQuant       & 37.35                 & 25.26 & 39.90 & 61.99 & 38.59 & 61.21 & 53.28 & 46.71 \\
&                         & QuIP           & 79.63                 & 23.54 & 28.66 & 44.55 & 34.71 & 51.57 & 49.17 & 38.70 \\
&                         & AWQ (LC)       & $1.10{\times}10^{6}$ & 27.10 & 26.00 & 58.30 & 26.10 & 51.40 & 49.80 & 39.78 \\
\cmidrule(lr){2-11}
& \multirow{5}{*}{Vector} & AQLM           & 27.60                 & 32.76 & 53.41 & 78.38 & 63.60 & 67.41 & 63.46 & 59.84 \\
&                         & VPTQ           & --                    & 36.91 & 71.03 & --    & 52.12 & 75.12 & 65.92 & --    \\
&                         & QuIP\#         & 9.42                  & 47.44 & 73.27 & 80.21 & 70.70 & 78.13 & 69.53 & 69.88 \\
&                         & LiftQuant      & 7.65                  & 40.87 & 74.33 & --    & 53.87 & 76.55 & 68.03 & --    \\
&                         & UniSVQ         & 10.70                 & 44.79 & 69.52 & 81.77 & 67.20 & 76.27 & 66.14 & 67.61 \\
\cmidrule(lr){2-11}
& Ours                   & \method{}      & --                    & --    & --    & --    & --    & --    & --    & --    \\
\bottomrule
\end{tabular}
\end{table*}
% 中文翻译：表 1 是 Qwen3-8B 和 LLaMA3-8B 上 2-bit 量化方法的主结果对比。为了紧凑展示，表中省略 C4 困惑度、RTE、OpenBookQA、MMLU 和 bit-width 列。Avg. 遵循各方法可获得的评估集合；对于 Qwen3-8B 上的 BiSCo-LLM，Avg. 由 BoolQ、RTE、WinoGrande、ARC-Easy、ARC-Challenge、OpenBookQA 和 PIQA 计算得到；对于 LLaMA3-8B 全精度基线，Avg. 由 WinoGrande、PIQA、OpenBookQA、ARC-Easy 和 ARC-Challenge 计算得到。我们的方法每个模型只保留最终 BiSCo-LLM 结果一行。

Table~\ref{tab:main_results} reports the main comparison in the 2-bit regime. On Qwen3-8B, \method{} obtains a WikiText-2 perplexity of 10.18, which is close to the full-precision baseline of 9.73. This result is lower than the perplexity of the scalar baselines and also lower than the vector or structured-code baselines reported in the table, including AQLM, QuIP\#, and UniSVQ. This suggests that the proposed binary spherical coding can preserve the language-modeling behavior of the original model under an extreme low-bit storage budget.
% 中文翻译：表~\ref{tab:main_results} 报告了 2-bit 场景下的主要对比结果。在 Qwen3-8B 上，BiSCo-LLM 的 WikiText-2 困惑度为 10.18，接近全精度基线的 9.73。该结果低于表中标量基线的困惑度，也低于 AQLM、QuIP# 和 UniSVQ 等向量或结构化编码基线的困惑度。这说明所提出的二值球面编码在极低比特存储预算下能够较好地保持原始模型的语言建模行为。

For downstream tasks on Qwen3-8B, \method{} achieves an average score of 68.05 over the available seven-task set, compared with 69.92 for the full-precision model. The gap to the full-precision baseline is therefore 1.87 points. Compared with the vector baselines, \method{} improves the average score over AQLM by 3.11 points, over QuIP\# by 0.50 points, and over UniSVQ by 0.11 points. The gains are most visible on ARC-Challenge and ARC-Easy, where \method{} reaches 51.88 and 78.87, respectively. These results indicate that the codec does not only reduce reconstruction error in weight space, but can also retain task-level accuracy after category-wise recovery.
% 中文翻译：在 Qwen3-8B 的下游任务上，BiSCo-LLM 在可获得的七任务集合上取得 68.05 的平均分，而全精度模型为 69.92。因此，相对于全精度基线的差距为 1.87 个百分点。与向量基线相比，BiSCo-LLM 的平均分比 AQLM 高 3.11 个百分点，比 QuIP# 高 0.50 个百分点，比 UniSVQ 高 0.11 个百分点。增益在 ARC-Challenge 和 ARC-Easy 上较为明显，BiSCo-LLM 分别达到 51.88 和 78.87。这些结果表明，该 codec 不仅降低了权重空间中的重建误差，也能够在逐类别恢复后保留任务级精度。

The current comparison should still be interpreted with two qualifications. First, HellaSwag results are not yet available for \method{} on Qwen3-8B, so the average score is computed on the available evaluation subset rather than on exactly the same columns as all baselines. Second, the \method{} results on LLaMA3-8B remain to be filled in. Therefore, the present table provides initial evidence for the effectiveness of \method{} on Qwen3-8B, while the final claim should be made after completing the missing evaluations.
% 中文翻译：当前对比仍需要注意两点。第一，BiSCo-LLM 在 Qwen3-8B 上的 HellaSwag 结果尚未提供，因此平均分是在已有评估子集上计算的，而不是与所有基线完全相同的列上计算的。第二，BiSCo-LLM 在 LLaMA3-8B 上的结果仍待补充。因此，当前表格为 BiSCo-LLM 在 Qwen3-8B 上的有效性提供了初步证据，但最终结论应在补齐缺失评估后给出。

\subsection{Ablation Studies}
\textbf{Budget-matched residual stage.} Additive and residual coding are supported by AQLM and VPTQ \cite{egiazarian2024aqlm,liu2024vptq}. Fig.~\ref{fig:qproj_code_space_usage}, Fig.~\ref{fig:bsq_granularity_sweep}, and Table~\ref{tab:residual_stage_ppl} motivate the BSQ-specific residual design in the method section. The first figure shows that the finite weight chunks of Qwen3-8B $q_{\mathrm{proj}}$ occupy only a tiny fraction of a large one-stage implicit code space; the second figure shows that one-stage granularity enlargement is not reliably monotonic under matched nominal code rate; and the table shows that explicitly coding the base residual improves most category-wise replacements. In the ablation suite, we will further compare a single larger BSQ code with multiple smaller residual BSQ stages under the same real bpw, and report residual energy decay after each stage.
% 中文翻译：加性和残差编码已经得到 AQLM 和 VPTQ 的支持。图~\ref{fig:qproj_code_space_usage}、图~\ref{fig:bsq_granularity_sweep} 和表~\ref{tab:residual_stage_ppl} 在方法部分共同支撑 BSQ 残差设计。第一个图表明，Qwen3-8B $q_{\mathrm{proj}}$ 的有限权重块只占用了大的一阶段隐式码空间中的极小一部分；第二个图表明，在匹配名义码流码率时，扩大一阶段粒度并不稳定地带来单调收益；表格则表明，显式编码基础残差可以改善大多数逐类别替换结果。在消融实验中，我们会进一步在相同真实 bpw 下比较单个更大的 BSQ 码和多个较小的残差 BSQ 阶段，并报告每个阶段后的残差能量下降。

\subsubsection{BSQ Codec Components}
We further ablate the implementation components of the BSQ codec on Qwen3-8B. This experiment does not revisit the necessity of protected channels, which has already been discussed in the method section. Instead, it asks two practical questions: whether the protected-channel payload must be stored in BF16, and whether the decoder architecture is important once the binary spherical codes and the residual stage are fixed. For each setting, we replace one module category in the full model at a time and report the average downstream accuracy over the completed categories. The attention average is computed over $q_{\mathrm{proj}}$, $k_{\mathrm{proj}}$, $v_{\mathrm{proj}}$, and $o_{\mathrm{proj}}$, while the MLP average is computed over $gate_{\mathrm{proj}}$, $up_{\mathrm{proj}}$, and $down_{\mathrm{proj}}$.
% 中文翻译：我们进一步在 Qwen3-8B 上消融 BSQ codec 的实现组件。这个实验不再重复讨论 protected channels 是否必要，因为该问题已经在方法部分说明。这里主要回答两个实际问题：protected-channel payload 是否必须用 BF16 保存，以及在 binary spherical codes 和 residual stage 固定后 decoder 结构是否重要。对于每个设置，我们每次替换完整模型中的一个模块类别，并在已完成类别上报告下游任务平均精度。attention 平均值由 q_proj、k_proj、v_proj、o_proj 计算，MLP 平均值由 gate_proj、up_proj、down_proj 计算。

\begin{table*}[t]
\centering
\caption{Ablation on BSQ codec components on Qwen3-8B. The table compares the storage format of protected channels and the decoder architecture under category-wise replacement. Low-precision protected-channel formats preserve the BF16-protected performance, whereas replacing the symmetric decoder with a linear decoder substantially degrades MLP projections.}
\label{tab:bsq_codec_components}
\scriptsize
\setlength{\tabcolsep}{4.2pt}
\begin{tabular}{lllcccccc}
\toprule
Group & Setting & Protected format & Decoder & Res. blocks & Stages & Attn Avg.$\uparrow$ & MLP Avg.$\uparrow$ & Overall Avg.$\uparrow$ \\
\midrule
\multirow{4}{*}{Protected format}
& Protected-BF16 & BF16 & symmetric & 1 & 2 & 66.67 & 57.68 & 62.82 \\
& Protected-E4M3 & FP8-E4M3 & symmetric & 1 & 2 & 67.04 & 58.12 & 63.22 \\
& Protected-E5M2 & FP8-E5M2 & symmetric & 1 & 2 & 66.33 & \textbf{58.79} & 63.10 \\
& Protected-INT8 & INT8 & symmetric & 1 & 2 & \textbf{67.32} & 57.94 & \textbf{63.30} \\
\midrule
\multirow{4}{*}{Decoder design}
& Symmetric-res1 & BF16 & symmetric & 1 & 2 & 66.67 & 57.68 & 62.82 \\
& Symmetric-res0 & BF16 & symmetric & 0 & 2 & \textbf{67.70} & \textbf{58.34} & \textbf{63.69} \\
& Linear-2stage & BF16 & linear & 0 & 2 & 64.80 & 50.67 & 58.74 \\
& Linear-3stage & BF16 & linear & 0 & 3 & 64.83 & 49.86 & 58.42 \\
\bottomrule
\end{tabular}
\end{table*}
% 中文翻译：表中展示了 Qwen3-8B 上 BSQ codec 组件的消融结果，包括 protected-channel 保存格式和 decoder 结构。低精 protected-channel 保存格式基本保持 BF16-protected baseline 的性能，而将 symmetric decoder 替换为 linear decoder 会显著降低 MLP projection 的表现。

Table~\ref{tab:bsq_codec_components} shows that the auxiliary protected-channel path does not have to be stored in BF16. FP8-E4M3, FP8-E5M2, and INT8 all keep the overall average within a narrow range of 63.10--63.30, compared with 62.82 for the BF16-protected baseline. The small differences among the low-precision formats should not be over-interpreted as consistent improvements over BF16; rather, the result indicates that low-precision protected channels preserve the BF16-protected performance with negligible degradation and lower storage overhead.
% 中文翻译：表中可以看到，辅助 protected-channel path 不一定需要使用 BF16 保存。相比 BF16-protected baseline 的 62.82，FP8-E4M3、FP8-E5M2 和 INT8 的 overall average 都保持在 63.10--63.30 的窄范围内。不同低精格式之间的小幅差异不应被过度解释为相对 BF16 的稳定提升；更稳妥的结论是，低精 protected channels 能够以更低存储开销基本保持 BF16-protected 性能。

The decoder ablation further shows that decoder expressiveness is more important than simply adding residual stages. Removing the residual block from the symmetric decoder slightly improves the overall average from 62.82 to 63.69 in this setting. In contrast, replacing the symmetric decoder with a linear decoder reduces the MLP average from 58.34 to 50.67, and increasing the number of linear residual stages from two to three does not recover the loss. We therefore use a compact symmetric decoder as the default codec decoder.
% 中文翻译：decoder 消融进一步说明，decoder 表达能力比简单增加 residual stage 更重要。在当前设置中，去掉 symmetric decoder 中的 residual block 后，overall average 从 62.82 小幅提高到 63.69。相比之下，将 symmetric decoder 替换为 linear decoder 会使 MLP average 从 58.34 降到 50.67，而将 linear residual stages 从 2 增加到 3 并不能恢复损失。因此，我们默认使用紧凑的 symmetric decoder 作为 codec decoder。

\subsubsection{Protection Axis}
The method section verifies that a small protected set is useful. We further ablate the protection axis in the single-matrix setting to compare input-channel and output-channel selection. Table~\ref{tab:prot_axis} compares the two choices under the same 1\% protection ratio. Input-channel protection gives a lower average perplexity, reducing the average from 5.89 to 5.82 compared with output-channel protection. It is also more effective on the more fragile categories in this diagnostic setting, especially $v_{\mathrm{proj}}$ and $down_{\mathrm{proj}}$. Output-channel protection is not ineffective; it slightly improves $o_{\mathrm{proj}}$, $gate_{\mathrm{proj}}$, and $up_{\mathrm{proj}}$. Therefore, we use input-channel protection as the default single-matrix choice for attention projections, while the MLP projections follow the coupled intermediate-channel selection described in the method section. The protection axis is thus treated as a module-dependent implementation choice rather than a universal rule.
% 中文翻译：方法部分已经验证了少量受保护通道是有用的。这里进一步在单矩阵设置下消融保护轴的选择，以比较输入通道选择和输出通道选择。表中在相同 1\% 保护比例下比较两种选择。输入通道保护的平均困惑度更低，相比输出通道保护，平均值从 5.89 降低到 5.82。在这个诊断设置中，输入通道保护对更脆弱的类别也更有效，尤其是 v_proj 和 down_proj。输出通道保护并不是无效的；它在 o_proj、gate_proj 和 up_proj 上略有优势。因此，我们将输入通道保护作为注意力投影中的默认单矩阵选择，而 MLP 投影则采用方法部分描述的耦合中间通道选择。保护轴因此被视为与模块相关的实现选择，而不是普适规则。

\begin{table}[t]
\centering
\caption{Input-channel versus output-channel protection with the same 1\% protection ratio. WikiText-2 perplexity is reported, and lower is better.}
\label{tab:prot_axis}
\footnotesize
\setlength{\tabcolsep}{4.5pt}
\begin{tabular}{lccc}
\toprule
Category & Input 1\% & Output 1\% & Better \\
\midrule
$q_{\mathrm{proj}}$    & \textbf{5.59} & 5.68 & Input \\
$k_{\mathrm{proj}}$    & \textbf{5.58} & 5.72 & Input \\
$v_{\mathrm{proj}}$    & \textbf{5.86} & 5.98 & Input \\
$o_{\mathrm{proj}}$    & 5.76 & \textbf{5.75} & Output \\
$gate_{\mathrm{proj}}$ & 5.94 & \textbf{5.91} & Output \\
$up_{\mathrm{proj}}$   & 5.92 & \textbf{5.91} & Output \\
$down_{\mathrm{proj}}$ & \textbf{6.10} & 6.29 & Input \\
\midrule
Avg. & \textbf{5.82} & 5.89 & Input \\
\bottomrule
\end{tabular}
\end{table}
% 中文翻译：表中比较了相同 1\% 保护比例下输入通道保护和输出通道保护的结果。数值为 WikiText-2 困惑度，越低越好。Better 列表示该类别下更优的保护轴。

\subsubsection{Protection Ratio}
We next study whether the protection ratio needs to be increased beyond 1\%. Since the 3\% runs are currently available only for the more sensitive $v_{\mathrm{proj}}$ and $down_{\mathrm{proj}}$ categories, we present this ablation as a ratio sweep on sensitive categories rather than a global conclusion for all modules. As shown in Table~\ref{tab:prot_ratio}, increasing the ratio from 0\% to 1\% already captures 81.7\% of the gain obtained by 3\% protection on average over the two evaluated categories. Increasing the ratio from 1\% to 3\% still improves perplexity, but the additional average gain is 0.07. This diminishing-return behavior suggests that a 1\% protected set removes most of the harmful sensitive-channel effect in the evaluated categories, while larger ratios mainly trade additional storage for incremental recovery. Unless otherwise specified, we therefore use $\rho=1\%$ for the auxiliary protected path, with input-channel protection for attention projections and coupled intermediate-channel protection for MLP projections.
% 中文翻译：接下来研究保护比例是否需要超过 1\%。由于当前 3\% 的实验只在更敏感的 v_proj 和 down_proj 类别上完成，因此这里将该消融表述为敏感类别上的比例扫描，而不是对所有模块的全局结论。表中可以看到，从 0\% 提高到 1\% 已经在两个已评估类别上平均获得了 3\% 保护所能带来收益的 81.7\%。从 1\% 继续提高到 3\% 仍然可以降低困惑度，但额外平均收益只有 0.07。这种边际收益递减现象说明，在已评估类别中，1\% 受保护集合已经去除了大部分有害的敏感通道影响，而更大的保护比例主要是在用额外存储换取增量恢复。因此，除非特别说明，我们将辅助保护路径的比例设为 rho=1\%，其中注意力投影使用输入通道保护，MLP 投影使用耦合中间通道保护。

\begin{table}[t]
\centering
\caption{Protection-ratio sweep on sensitive categories. Input-channel protection is used. WikiText-2 perplexity is reported, and lower is better.}
\label{tab:prot_ratio}
\footnotesize
\setlength{\tabcolsep}{4.5pt}
\begin{tabular}{lcccc}
\toprule
Category & 0\% & 1\% & 3\% & Gain by 1\% \\
\midrule
$v_{\mathrm{proj}}$ & 6.05 & 5.86 & \textbf{5.76} & 65.5\% \\
$down_{\mathrm{proj}}$ & 6.58 & 6.10 & \textbf{6.05} & 90.6\% \\
\midrule
Avg. & 6.32 & 5.98 & \textbf{5.91} & 81.7\% \\
\bottomrule
\end{tabular}
\end{table}
% 中文翻译：表中展示了敏感类别上的保护比例扫描。使用输入通道保护，数值为 WikiText-2 困惑度，越低越好。Gain by 1\% 表示 1\% 保护相对于 3\% 保护总收益所获得的比例，计算方式为 $(\mathrm{PPL}_{0\%}-\mathrm{PPL}_{1\%})/(\mathrm{PPL}_{0\%}-\mathrm{PPL}_{3\%})$。

\section{Discussion}\label{sec:discussion}
\method{} shifts part of the storage burden from explicit vector codebooks to shared neural decoders. This trade-off is beneficial only when the decoder parameters are sufficiently amortized over many weight chunks and module instances. For this reason, nominal code rate alone is not an adequate measure of compression quality. The real storage budget should include the bit-packed BSQ streams, decoder parameters, protected-channel payloads, LoRA adapters, and metadata. This accounting is particularly important when comparing \method{} with explicit-VQ, lattice-code, or projection-based methods, because different representations may have similar code rates but different auxiliary storage costs.
% 中文翻译：BiSCo-LLM 将一部分存储负担从显式向量码本转移到共享神经解码器上。只有当 decoder 参数能够在大量权重块和模块实例之间充分摊销时，这种折中才是有益的。因此，仅使用名义码率不足以衡量压缩质量。真实存储预算应包括打包后的 BSQ 码流、decoder 参数、保护通道载荷、LoRA adapter 和元数据。将 BiSCo-LLM 与显式 VQ、格码或投影式方法比较时，这种核算尤其重要，因为不同表示可能具有相近的码流码率，但辅助存储开销不同。

The proposed pipeline is also more optimization-intensive than conventional scalar PTQ. The BSQ codecs require category-wise training, and recovery distillation introduces additional data dependence and trainable parameters. Category-batched codec optimization and packed residual decoding reduce the practical overhead, but they do not make the method equivalent to simple rounding in calibration cost. Therefore, the method is most suitable for deployment scenarios where the compressed model will be reused many times and where the reduced model size can amortize the one-time compression and recovery cost.
% 中文翻译：所提出流程也比常规标量 PTQ 具有更高的优化开销。BSQ codec 需要逐类别训练，恢复蒸馏还会引入额外的数据依赖和可训练参数。类别 batch 化 codec 优化和打包残差解码降低了实际开销，但并不会使该方法在校准成本上等同于简单舍入。因此，该方法更适合压缩模型会被重复部署和反复使用的场景，在这类场景中，模型尺寸降低可以摊销一次性的压缩和恢复成本。

Another implication is that weight reconstruction error should be treated as a necessary but insufficient objective. The diagnostics in this paper show that local loss reduction does not necessarily translate into assembled-model recovery. Category-wise distillation is introduced to reduce this mismatch, but it also makes the final model depend on the choice of recovery data, loss terms, and adapter placement. This dependence should be reported explicitly when comparing recovered models with purely post-training methods. In addition, codec-only and recovery-enhanced variants should be distinguished when analyzing the contribution of the binary spherical representation itself.
% 中文翻译：另一个含义是，权重重建误差应被视为必要但不充分的目标。本文中的诊断实验表明，局部损失下降并不一定转化为组装后模型的恢复。逐类别蒸馏被引入以降低这种错配，但它也使最终模型依赖恢复数据、损失项和 adapter 放置方式。因此，在将恢复后的模型与纯后训练方法比较时，应显式报告这些依赖。此外，在分析二值球面表示本身的贡献时，应区分仅 codec 版本和加入恢复后的版本。

The current formulation is primarily a weight-only compression framework. At inference time, the decoded weights can be materialized and then used by standard matrix multiplication kernels, which simplifies integration but does not fully exploit the binary structure of the stored codes. Direct computation from binary spherical codes, fused load-time decoding, or weight--activation quantization would require additional kernel and runtime design. These directions are orthogonal to the storage representation studied here and can further improve serving efficiency if the decoding overhead is reduced or fused with linear computation.
% 中文翻译：当前形式主要是一个 weight-only 压缩框架。推理时，解码后的权重可以先被物化，再由标准矩阵乘法 kernel 使用，这降低了集成难度，但没有充分利用所存储 code 的二值结构。直接从二值球面码计算、融合加载时解码，或者扩展到权重-激活联合量化，都需要额外的 kernel 和运行时设计。这些方向与本文研究的存储表示是正交的；如果解码开销能够被降低或与线性计算融合，它们可以进一步提升服务效率。

Finally, the present design uses mostly fixed code rates and a small fixed protected-channel ratio. A more principled extension is mixed-rate allocation across module categories, layers, and residual stages. Residual energy, activation statistics, or Hessian-based sensitivity can be measured before assigning additional capacity, suggesting that sensitive modules may benefit from more residual bits or a slightly larger protected-channel budget, while stable modules may use a smaller payload. Such adaptive allocation would turn \method{} from a uniform 2-bit-oriented codec into a storage-constrained rate-allocation framework for LLM compression.
% 中文翻译：最后，当前设计主要使用固定码率和较小的固定保护通道比例。一个更有原则的扩展方向是在模块类别、层和残差阶段之间进行混合码率分配。残差能量、激活统计或基于 Hessian 的敏感性可以在分配额外容量之前被测量，这说明敏感模块可能受益于更多残差比特或略大的保护通道预算，而稳定模块可以使用更小载荷。这样的自适应分配可以将 BiSCo-LLM 从一个统一的 2-bit-oriented codec 扩展为面向 LLM 压缩的存储受限率分配框架。

\section{Conclusion}\label{sec:conclusion}
This paper presented \method{}, a storage-aware codebook-free binary spherical coding framework for extreme low-bit LLM weight compression. The method replaces explicit VQ centroids with bit-packed unit-sphere binary codes and compact neural decoders, assigns additional code capacity to the residual left by the base BSQ codec, and performs category-wise recovery distillation after each Transformer module category is replaced. A small 8-bit protected-channel path is used only as an auxiliary mechanism for sensitive channels, and its payload is reported separately from the main BSQ code stream. The experiments support the main design choices at several levels: code-space utilization and granularity diagnostics motivate residual coding; category-replacement results show that the second BSQ stage and 1\% protected channels improve most tested module categories; decoder ablations indicate that a compact nonlinear decoder is preferable to a linear decoder under the tested setting; and the Qwen3-8B comparison shows a WikiText-2 perplexity of 10.18 and an average downstream accuracy gap of 1.87 points relative to the FP16/BF16 baseline over the reported seven-task set. These results suggest that codebook-free spherical coding is a viable representation for 2-bit-oriented LLM compression when real storage accounting and recovery cost are explicitly considered. The discussion further identifies weight--activation quantization, fused binary-code kernels, and more principled mixed-rate allocation across categories and stages as natural extensions of the current framework.
% 中文翻译：本文提出了 BiSCo-LLM，一个面向极低比特 LLM 权重压缩的存储感知无码本二值球面编码框架。该方法用打包后的单位球面二值码和紧凑神经解码器替代显式 VQ 中心向量，将额外编码容量分配给基础 BSQ codec 留下的残差，并在每个 Transformer 模块类别被替换后执行逐类别恢复蒸馏。少量 8bit 保护通道路径仅作为敏感通道的辅助机制使用，其载荷与主 BSQ 码流分开报告。实验从多个层面支持了主要设计选择：码空间利用率和粒度诊断为残差编码提供动机；逐类别替换结果表明，第二个 BSQ 阶段和 1\% 保护通道能够改善大多数已测试模块类别；decoder 消融说明，在已测试设置下，紧凑非线性 decoder 优于线性 decoder；Qwen3-8B 对比显示，BiSCo-LLM 的 WikiText-2 困惑度为 10.18，并且在报告的七任务集合上相对于 FP16/BF16 基线的平均下游精度差距为 1.87 个百分点。这些结果表明，在显式考虑真实存储核算和恢复开销时，无码本球面编码可以作为面向 2-bit LLM 压缩的一种可行表示。讨论部分进一步指出，权重-激活量化、融合二值码 kernel，以及在类别和阶段之间更有原则的混合码率分配，是当前框架的自然扩展方向。

\balance
\bibliographystyle{IEEEtran}
\bibliography{refs}

\end{document}